\documentclass[lettersize,journal]{IEEEtran}
\usepackage{amsmath,amsfonts}
\usepackage{algorithmic}
\usepackage{algorithm}
\usepackage{array}
\usepackage[caption=false,font=normalsize,labelfont=sf,textfont=sf]{subfig}
\usepackage{textcomp}
\usepackage{stfloats}
\usepackage{url}
\usepackage{verbatim}
\usepackage{graphicx}
\usepackage{cite}
\usepackage{textcomp}
\usepackage{makecell}
\usepackage{rotating}
\usepackage{url}

\newcommand{\etal}{\textit{et al.}}
\begin{document}

\title{ConvShareViT: Enhancing Vision Transformers with Convolutional Attention Mechanisms for Free-Space Optical Accelerators}

\author{

Riad Ibadulla,
Thomas M. Chen, and
Constantino Carlos Reyes-Aldasoro
        % <-this % stops a space
\thanks{All authors are with School of Science and Technology, City St. George’s, University of London, EC1V
0HB, London, UK}% <-this % stops a space
\thanks{Constantino Carlos Reyes-Aldasoro is also with the Integrated Pathology Unit, Institute of Cancer Research, Sutton, UK}}
% The paper headers

% \IEEEpubid{0000--0000/00\$00.00~\copyright~2021 IEEE}
% Remember, if you use this you must call \IEEEpubidadjcol in the second
% column for its text to clear the IEEEpubid mark.

\maketitle

\begin{abstract}
This paper introduces ConvShareViT, a novel deep learning architecture that adapts Vision Transformers (ViTs) to the 4f free-space optical system. ConvShareViT  replaces linear layers in multi-head self-attention (MHSA) and Multilayer Perceptrons (MLPs) with a depthwise convolutional layer with shared weights across input channels. Through the development of ConvShareViT, the behaviour of convolutions within MHSA and their effectiveness in learning the attention mechanism were analysed systematically. Experimental results demonstrate that certain configurations, particularly those using valid-padded shared convolutions, can successfully learn attention, achieving comparable attention scores to those obtained with standard ViTs. However, other configurations, such as those using same-padded convolutions, show limitations in attention learning and operate like regular CNNs rather than transformer models.
ConvShareViT architectures are specifically optimised for the 4f optical system, which takes advantage of the parallelism and high-resolution capabilities of optical systems. Results demonstrate that ConvShareViT can theoretically achieve up to 3.04 times faster inference than GPU-based systems. 
% This potential acceleration makes ConvShareViT an attractive candidate for future optical deep learning applications and proves that our ViT (ConvShareViT) can be employed using only the convolution operation, though further optimisation is necessary to balance performance and complexity.
This potential acceleration makes ConvShareViT an attractive candidate for future optical deep learning applications and proves that our ViT (ConvShareViT) can be employed using only the convolution operation, via the necessary optimisation of the ViT to balance performance and complexity.
\end{abstract}

\section{Introduction}

Computer vision has advanced significantly with the development of the deep learning approaches~\cite{krizhevsky_imagenet_2012,wang_yolov10_2024,he_deep_2016,ronneberger_u-net_2015,dosovitskiy_image_2021}. Convolutional Neural Networks (CNNs) have been the standard approach for tasks like image classification~\cite{krizhevsky_imagenet_2012,deng_imagenet_2009}, object detection~\cite{wang_yolov10_2024}, and image segmentation~\cite{badrinarayanan_segnet_2017,ronneberger_u-net_2015,ibadulla_fat-u-net_2024,karabag_impact_2023} for a long time due to their ability to efficiently capture spatial hierarchies in images~\cite{krizhevsky_imagenet_2012}. CNNs were challenged by the invention of Vision Transformers (ViT)~\cite{dosovitskiy_image_2021}, which use the encoder of the transformer and its self-attention mechanisms~\cite{vaswani_attention_2017}. Unlike the CNNs that treat an image as a whole, ViTs divide the images into patches, treating each patch as a sequence token. This approach has shown outstanding success \cite{Parvaiz2023}, when models pre-trained on large datasets like ImageNet are fine-tuned on smaller, task-specific datasets. However, CNNs still stay the first choice for small datasets when trained from scratch, due to their simplicity and sometimes better performance over transformers for specific datasets~\cite{gai_comparing_2024,ascencio-cabral_comparison_2022}. 

As deep learning models grow in complexity, advanced hardware accelerators \cite{Armeniakos2022} are increasingly needed to efficiently manage their computational demands. Conventional electronic accelerators struggle with limitations in power consumption and processing speed as model sizes continue to expand \cite{Peccerillo2022}. Optical computing \cite{Sui2020,Kazanskiy2022}, particularly the 4f system, offers a promising solution due to its ability to process data rapidly and with high energy efficiency, taking advantage of light’s parallel processing capabilities. A 4f system, first described by Weaver and Goodman in 1966~\cite{weaver_technique_1966}, is an optical setup consisting of a laser, modulators, lenses, and a camera or photodetector, with each component separated by one focal length of the lens. The entire arrangement spans four focal lengths, hence the name `4f system'. This system, also known as a 4f correlator, is commonly used to perform the convolution operation in optics. %It was first described by Weaver and Goodman in 1966~\cite{weaver_technique_1966}.

The 4f system is particularly effective for accelerating CNNs through its efficient handling of convolution operations. While various optical setups have been developed for specific neural networks~\cite{li_channel_2020, chang_hybrid_2018, miscuglio_massively_2020,colburn_optical_2019,gupta_4f_2022,ibadulla_fatnet_2023,dai_-chip_2023,schultz_optical_2021,chen_multilayer_2023}, the 4f system's ability to accelerate CNNs could, in theory, be generalised across different network architectures, enabling a single device to support multiple model types. This creates the potential for a standardised optical approach to training neural networks.

With ViTs gaining popularity in computer vision, using convolutional layers within the multi-head self-attention (MHSA) layers on the same 4f system could simplify the hardware landscape. This eliminates the need for multiple specialised processors, streamlining both training and inference and making optical computing more practical and scalable for complex neural network tasks.

% As deep learning models grow in complexity, they increasingly require advanced hardware accelerators to meet their computational demands efficiently. Conventional electronic accelerators face limitations in power consumption and processing speed as model sizes expand. Optical computing, particularly using the 4f system, provides a promising alternative due to its rapid processing capabilities and high energy efficiency. The 4f optical system excels in performing convolution operations quickly by taking advantage of the high-speed, parallel processing of light. Unfortunately, when it coems to the optical acceleration of neural networks, the researchers been building different setups for particular types of the neural networks. Among all, the use of the 4f system for performing the convolution operation in the 4f system to accelerate the CNN is the standard approach, which in theory could have been applied to all range of neural networks. In the work of Ibadulla~\etal the authors describe the method of converting any CNN into the format most optimal for the 4f optical training. Since the ViTs gaining the popularity in the computer vision field, running both convolutional layers and multi-head self-attention (MHSA) layers in the same 4f device simplifies the hardware landscape, eliminating the need for multiple specialised processors. This unified approach not only streamlines the training and inference processes but also makes optical computing more practical and scalable for complex neural network tasks in computer vision and beyond.

In this paper, we propose the Convolutional Shared Vision Transformers (ConvShareViT), a hybrid model that incorporates convolutional operations within the Vision Transformer architecture, enhancing its adaptability to optical systems like the 4f. This model strictly uses convolutional operations in the MHSA layers and replaces traditional MLPs with convolution-based structures, ensuring compatibility with the 4f system. We analysed strategies such as channel/kernel/mix tiling  to optimise these operations and efficiently manage the computational load. Moreover, we analyse different methods of using convolution operations within the ViT's MHSA layers.

Unlike other hybrid models such as CvT~\cite{wu_cvt_2021}, which alter the intrinsic structure of Vision Transformers by incorporating pooling and downsizing mechanisms, our ConvShareViT strictly preserves the patch-based processing fundamental to the Vision Transformer philosophy. This ensures that each image patch is processed independently through convolution-enhanced self-attention layers, respecting the original design while improving the model’s ability to generalise from limited data by leveraging the locality and invariance properties typical of CNNs. While our work explores convolutional strategies similar to CvT, one of our objectives is also to demonstrate that MHSA can be implemented within a 4f optical system. This capability is essential not only for image classification, but also for enabling a wider range of transformer-based applications in optical computing. Therefore, we prioritise the replication and extension of the original ViT architecture, rather than adopting structural changes characteristic of CvT.

Our contributions are significant in two key areas: firstly, we demonstrate that convolutional operations can be seamlessly integrated into Vision Transformers without compromising their essential patch-processing methodology. Secondly, we demonstrated the capability of ConvShareViT to learn attention through a systematic evaluation and visualisation of our models' performance with different methods of incorporating convolution operations into the MHSA and MLP of the ViT.

% The subsequent sections will delve into the architectural design of the Vision Conformer, its implementation on the 4F system, and a detailed comparison of its performance against baseline models across a range of benchmarks. Through this exploration, we aim to bridge the gap between the theoretical appeal of Vision Transformers and their practical applicability, especially in scenarios where data and computational resources are limited.

% Computer vision, which can involve a large number of images with very slight differences, is considered to be one of the most complex problem areas for AI. Within the deep learning approaches, convolutional neural networks (CNNs) have become a standard approach for various computer vision problems.

\section{Related work}
Deep learning in computer vision has advanced significantly with the development of CNNs, and the introduction of Vision Transformers has elevated the field to new heights~\cite{patel_upsurge_2020}. While both approaches are much more efficient than fully connected neural networks, the need to accelerate Deep Learning performance has become increasingly critical. Various techniques have been employed to speed up deep learning, such as using shallower networks~\cite{ba_deep_2014}, pruning redundant weights~\cite{han_deep_2016}, or adopting lower quantisation levels~\cite{rastegari_xnor-net_2016}. Moreover, hardware accelerators, such as application-specific integrated circuits (ASICs), have been used to greatly enhance training and inference speeds, often outperforming conventional CPUs/GPUs.

However, as Moore's law slows down~\cite{waldrop_chips_2016} and the limits of electronic hardware become clearer, the need for alternative methods, such as optical accelerators, is growing more important. Optical computing, with its ability to process many tasks in parallel and at high speed, provides a promising way to overcome the challenges of traditional hardware and improve the performance of deep learning systems.

There are two main approaches to optical neural networks: the free-space method, and the silicon photonics approach. The free-space method involves the propagation of light through mediums such as air or  vacuum, without the need for physical waveguides and uses spatial light modulators (SLMs)~\cite{schultz_optical_2021,dai_-chip_2023,li_channel_2020,gupta_4f_2022,miscuglio_massively_2020,chen_multilayer_2023}. The silicon photonics approach uses Mach–Zehnder interferometers (MZIs)~\cite{hughes_training_2018,shen_deep_2017}, which  offer faster processing speeds, with clock rates reaching several GHz. However, it offers inferior parallelism compared to the free-space system.

%the free-space method, which uses spatial light modulators (SLMs)~\cite{schultz_optical_2021,dai_-chip_2023,li_channel_2020,gupta_4f_2022,miscuglio_massively_2020,chen_multilayer_2023}, and the silicon photonics approach, which uses Mach–Zehnder interferometers (MZIs)~\cite{hughes_training_2018,shen_deep_2017}. Unlike silicon photonics, free-space involves the propagation of light through mediums such as air or a vacuum, without the need for physical waveguides. Although the silicon photonics approach offers faster processing speeds, with clock rates reaching several GHz, it offers inferior parallelism compared to the free-space system.
The reader is referred to~\cite{sui_review_2020}~for a comprehensive review of different methods of implementing optical neural networks, comparing free-space techniques with silicon photonics methods, the latter being described as waveguide optical interconnection in their work.

Although direct optical matrix-vector multiplication approaches, such as demonstrated by Anderson~\etal~\cite{anderson_optical_2023}, implement Transformer architectures using free-space optics, our ConvShareViT model contributes uniquely by showing how existing 4f optical systems—already widespread for convolution operations—can be directly used to implement Vision Transformers. Our findings are significant because they indicate not only a path for a broad adoption of optical Transformers within existing devices but also deepen our theoretical understanding by proving that shared depthwise convolution is capable of learning attention. Therefore, our approach complements specialised MVM accelerators by expanding the versatility of convolution-based optical computing.

\subsection{Free-space optical image classification}

\begin{figure}[h]
    \centering
    \includegraphics[width=0.98\linewidth]{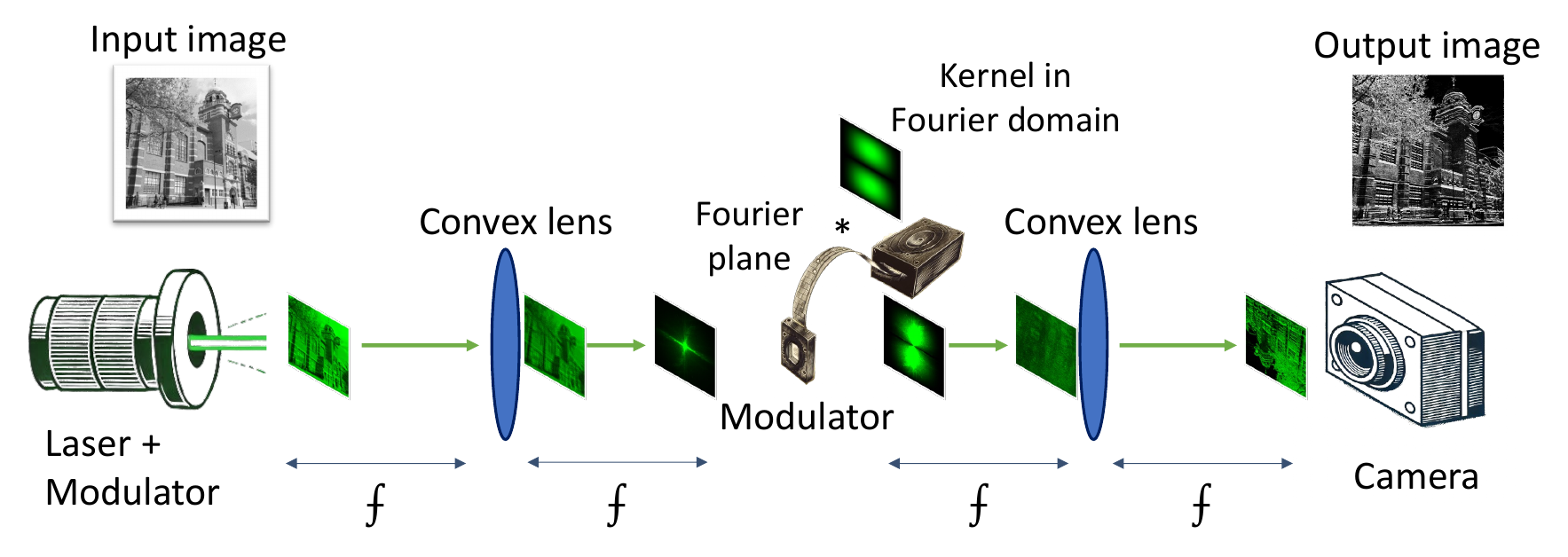}
    \caption{Schematic illustration of a 4f optical system executing a convolution operation. The system consists of an input plane (laser source), a convex lens, and a Fourier plane (containing a modulator or phase mask), followed by another convex lens and a camera, each positioned one focal length away from the lenses. As the light passes through the first lens, it undergoes a 2D Fourier transform at the Fourier plane, where it is multiplied by the kernel in the frequency domain. The light then travels through the second lens, which transforms it back to the spatial domain, and the camera captures the output.}
    \label{fig:schedamtic_4f}
\end{figure}

In this study, we focus on active 4f free-space optical accelerators, which play a crucial role in high-speed image processing and classification tasks. The 4f system is an optical setup used primarily for performing convolutions, a core operation in computer vision tasks such as image classification. The 4f system takes its name from the optical configuration, in which a light beam travels through four focal lengths between lenses and spatial modulators to perform a Fourier transform and its inverse as illustrated in Fig.~\ref{fig:schedamtic_4f}. This system's ability to perform optical convolution exploits the Fourier transform, where the convolution operation in the spatial domain becomes a simple pointwise multiplication in the frequency domain.

The process of the convolution operation in the 4f system with the simulated results is shown in Fig.~\ref{fig:schedamtic_4f} and performed as follows: First, light is directed onto the light modulator, where it is modulated by the input image. When the beam then passes through a convex lens, the resulting wavefront, after travelling one focal distance, represents the Fourier Transform of the image. Another light modulator, positioned in the Fourier plane, is used to perform element-wise multiplication with the image in the frequency domain. The modified beam then passes through a second lens, which performs the inverse Fourier Transform, converting the image back into the spatial domain. The final output, after another focal distance, is the convolved image, which is captured by a camera.

\subsection{Parallelism in 4f system}

% The primary advantage of 4f free-space optical neural networks over the on-chip silicon photonics method is that the 4f system enables massive parallelism, allowing for higher-resolution inferences and multiple simultaneous inferences. 

The primary advantage of 4f free-space optical neural networks is their ability to enable massive parallelism, allowing high-resolution inferences and the execution of multiple inferences simultaneously. This allows the 4f system to effectively process high-resolution inputs and kernels without compromising the frame rate. Since the 4f system efficiently processes high-resolution inputs, it's possible to parallelise the inference through the 4f system by tiling the inputs into a large block of inputs and performing the convolution of several inputs with the same kernel. This method is known as input tiling or batch tiling~\cite{ibadulla_fatnet_2023}, as the inputs of the entire mini-batch can be tiled to fit the entire input tensor~(n-dimensional matrix) of the CNN.

Since the convolution operation is commutative, the order of the operands does not matter. This means that tiling can be applied to the kernels instead of inputs, given that the input is a single 2D feature map. This approach has been used by Chang ~\etal~\cite{chang_hybrid_2018} and also described by Li~\etal~\cite{li_4f_2021} as kernel tiling, where the kernels are zero-padded to a size of $ (M+N-1) \times (M+N-1) $, with $ M \times M $ being the input resolution and $ N \times N $ representing the kernel resolution. After padding, these kernels are arranged into a single large tiled kernel block. The input must also be padded to match the dimensions of this kernel block to enable optical convolution. The resulting output consists of multiple tiled sections, each corresponding to the convolution of the input feature map with one of the kernels from the tiled array.

Another approach by Li~\etal~\cite{li_channel_2020}~achieved parallelism in CNNs using the 4f system, which is channel tiling. Similar to kernel tiling, this method involves padding of both kernels and channels. The padded channels and kernels are then tiled into respective blocks and convolved. The result of the convolution produces a block of outputs with dimensions $(2\sqrt{N_c}-1) \times (2\sqrt{N_c}-1)$, where $N_c$ is the number of input channels. All outputs, except for the one in the centre, are considered invalid and discarded. The valid output represents the sum of the convolution of each input channel with its corresponding kernel. As this method provides both the convolution and the summation of the results, it can be utilised in the channel summation process.

Unfortunately, this method computes only a single output channel. The third and most efficient method, as described by Li~\etal, is mixed tiling, which allows the entire convolutional layer to be performed in a single inference through the 4f system. Mixed tiling combines both kernel tiling and channel tiling, providing full parallelism for the entire convolutional layer. This method ensures that the convolution of all input channels with their respective kernels is completed, and the results are summed across the output channels. However, it requires a significant amount of spatial space within the 4f system, often making it impractical to execute the entire process in a single inference.

In this method, inputs are padded as before and tiled horizontally. Likewise, the kernels are also padded to the same dimensions and tiled along both the $x$ and $y$ axes, with each row corresponding to an output channel. Similar to channel tiling, the output block contains invalid regions due to unnecessary convolutions, while the valid outputs are located in the centre of each row of the output block.

\subsection{Vision Transformers}

ViTs represent a paradigm shift in the field of computer vision, achieving state-of-the-art results by using transformer-based architectures, initially designed for natural language processing (NLP) \cite{Han2023,Liu2024}. Unlike traditional CNNs, which rely on convolutional operations to extract hierarchical features from images, ViTs apply the transformer’s attention mechanism to capture global dependencies between different regions of an image. This is illustrated in %can be seen from 
the top half of Fig.~\ref{fig:Compparison_Vit_ConvShareViT}. 

At the core of ViTs is the self-attention mechanism, a component originally introduced in the transformer models for NLP. The attention mechanism allows the network to weigh the relationships between different parts of the input data, enabling it to capture contextual information across the entire image. To process an image, the ViT model divides it into fixed-size patches, which are treated similarly to tokens in NLP models (Fig.~\ref{fig:Compparison_Vit_ConvShareViT}). Each patch is then flattened into a vector, and positional embeddings are added to retain spatial information, ensuring that the ViT is aware of the position of the patches in the original image. This helps to maintain the spatial structure which could be lost after the tokenisation.

The tokens are passed through consecutive MHSA and MLP layers and repeated $N$ times. The depth of the transformer is a crucial hyper-parameter, which depends on the task. 

\begin{figure}
    \centering
    \includegraphics[width=0.98\linewidth]{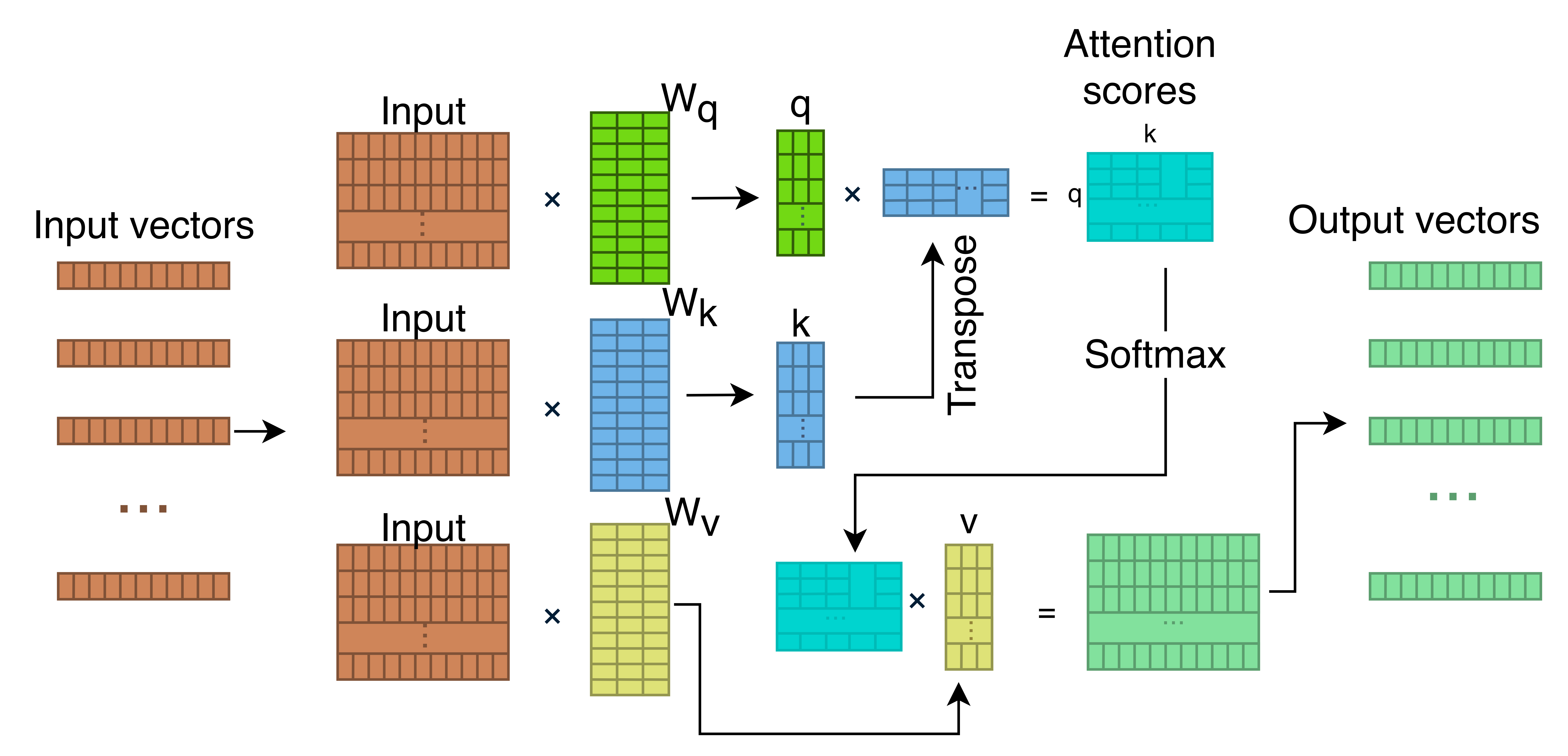}
    \caption{Self-Attention mechanism. Inputs are mapped to Query, Key, and Value vectors. Attention scores, calculated from Query-Key multiplications, achieving dependencies between tokens. These scores are then used to weight the Value matrix, amplifying relevant information.}
    \label{fig:normal_mhsa_diagram}
\end{figure}

The pipeline for the self-attention mechanism is shown in Fig.~\ref{fig:normal_mhsa_diagram}, where the tokens are part of the matrix X. The input matrix is multiplied by matrices $W_q$, $W_k$, and $W_v$ produce new sets of $Q$ (Query), $K$ (Key) and $V$ (Value) matrices. The multiplication of the input matrix by $W_q$, $W_k$, and $W_v$ is simply achieved using a linear layer without a bias term. It is important to note that each row of the input matrix (token) is mapped to the $Q$, $K$, and $V$ independently without any interaction with the adjacent tokens. The interaction between the tokens occurs later when calculating the attention scores through matrix multiplication $QK^\top$. 

The self-attention of each head is computed as follows:
\begin{equation}
    Attention(Q,K,V) = softmax(\frac{QK^\top}{\sqrt{d}})V,
    \label{eq:attention_matrix}
\end{equation}
where $d$ is the dimensionality of the $QKV$ vectors, included to scale the results of attention scores and improve numerical stability in the softmax calculation. $Q$, $K$, and $V$ represent query, key, and value matrices respectively, which are mapped from the input embeddings by linear projections. 

Multi-Heads Self-Attention (MHSA) layer processes multiple self-attention mechanisms in parallel, each focusing on different aspects of the patches. 

\subsection{Use of convolution operations in transformers}

One of the ViT variations, which used a similar shape of the network as the CNNs, is the Pyramid Vision Transformer (PVT)\cite{wang_pyramid_2021}. PVT reshapes feature maps into a matrix form after each transformer block, creating a pyramid-like architecture where the feature map size progressively reduces. This hierarchical structure similar to the CNNs, unlike the standard ViT, is better suited for visual tasks such as classification. Notably, PVT-Large delivers comparable top-1 ImageNet accuracy (a 0.1\% decrease) to ViT-Base/16~\cite{dosovitskiy_image_2021} while significantly reducing computational complexity (9.8 GFLOPs vs. 17.6 GFLOPs) and the number of parameters (61.4M vs. 86.6M).

While PVT mimics CNN-like feature extraction through its cone-shaped architecture, it does not incorporate convolutional layers. However, another model known as the Convolutional Vision Transformer (CvT)~\cite{wu_cvt_2021} model introduces convolution operations into the ViT pipeline, where overlapping convolutions are used for token embedding. This step captures local spatial information while simultaneously reducing the sequence length and increasing token feature dimensions, similar to CNN architectures. The convolutional token embedding stage also spatially down-samples the feature maps while increasing the number of channels, which enhances the model’s hierarchical representation capabilities.

In each Convolutional Transformer Block, CvT replaces the standard matrix multiplication for query, key, and value embeddings with depthwise separable convolutions. After each transformer block, the 2D feature map is reconstructed and passed through another convolutional token embedding layer, ensuring a cone-shaped architecture similar to CNNs used in classification tasks.

Unlike standard ViTs, which process non-overlapping patches and rely solely on the attention mechanism for communication between patches, CvT re-tokenizes the feature map after each transformer block, allowing for integrated feature representations. Additionally, due to the use of overlapping patches, CvT eliminates the need for positional encoding. 

% Wu \etal~\cite{wu_convolutional_2021} demonstrated that omitting positional encoding led to a slight improvement in ImageNet top-1 accuracy (from 81.4\% to 81.6\%).

In the work of Ding~\etal~\cite{ding_scaling_2022}, the authors revisit large kernel design in convolutional networks, proposing RepLKNet, a CNN architecture with kernel sizes as large as $31 \times 31$. This work is particularly notable because, although it does not employ the attention mechanisms seen in ViTs, it effectively closes the performance gap between CNNs and ViTs. According to the authors, the reason for the strong performance in ViTs is their high effective receptive field (ERF), which can be replicated in CNNs by leveraging large depthwise convolutions. This allows the model to capture both local and global information, similar to what attention mechanisms achieve in ViTs. Ding~\etal~demonstrate that the large kernels of RepLKNet deliver competitive results on ImageNet, achieving 84.8\% top-1 accuracy, which is on par with the Swin Transformer~\cite{liu_swin_2021} but with lower latency. Their approach shows that CNNs, through the strategic use of large kernels, can match or even surpass ViT performance without relying on attention mechanisms.

\section{Proposed method}

Regular vision Transformers can be divided into four main components: tokenisation, multi-head self-attention, multilayer perceptron, and classifier head. Multi-head self-attention, on the other hand, can be split into three main stages: $QKV$ projection, attention score calculation, and weighted sum of values. In this section, we will describe how each task is transformed to use only convolution operations.

\begin{figure*}[!htbp]
    \centering
    \includegraphics[width=\linewidth]{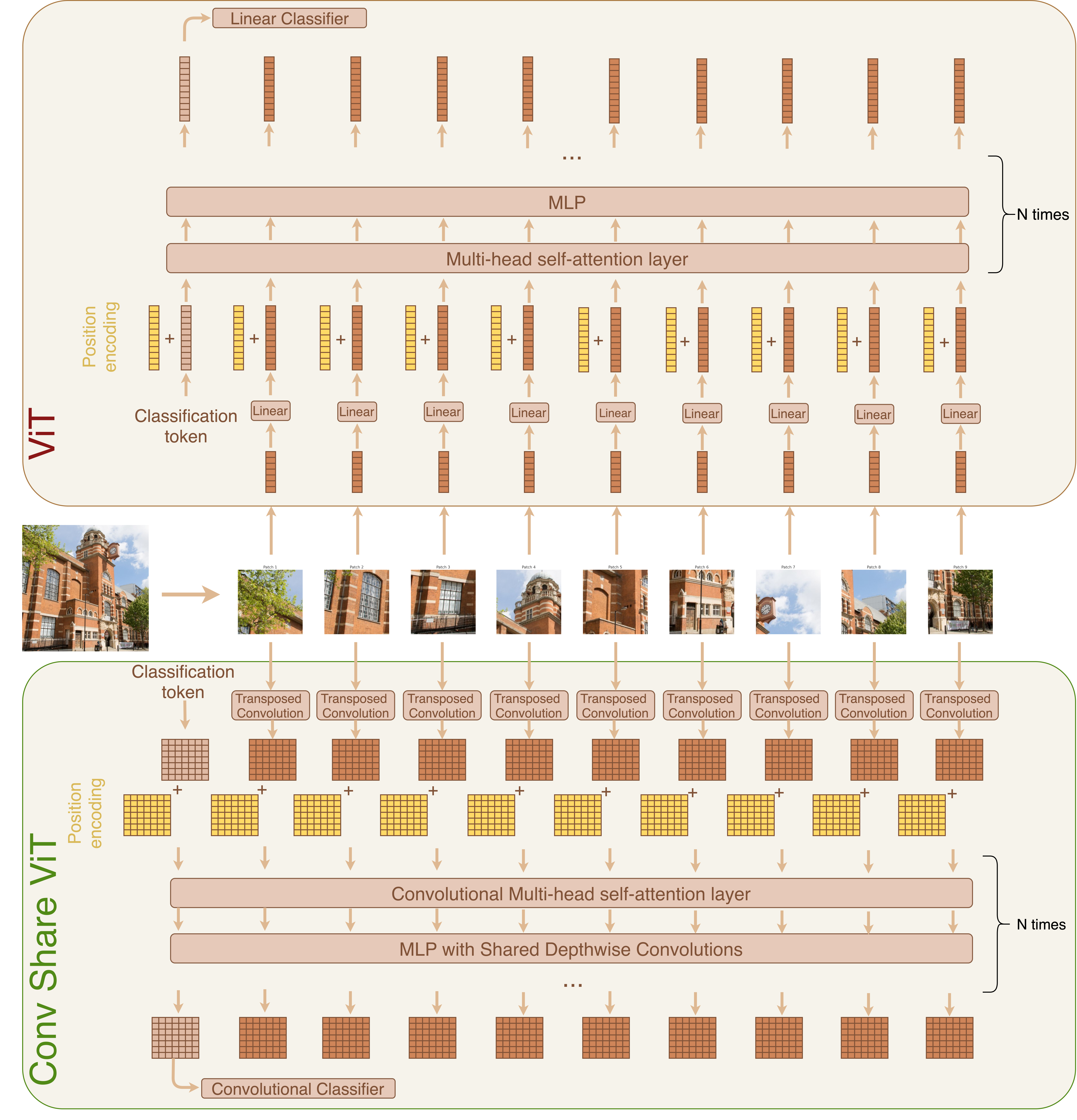}
    \caption{Comparison of the regular ViT (top of figure) and ConvShareViT (bottom of the figure) pipelines. ViTs vectorise patches of the image and apply a linear layer to map them into higher dimensional embeddings, while ConvShareViT keeps the patches in 2D format and uses Transpose Convolution to increase the dimensionality. ConvShareViT uses MHSA and MLPs using Shared Depthwise Convolutional layers.}
    \label{fig:Compparison_Vit_ConvShareViT}
\end{figure*}

\subsection{Shared depthwise convolution}
\begin{figure}[h]
    \centering
    \begin{tabular}{l}
         \includegraphics[width=0.9\linewidth]{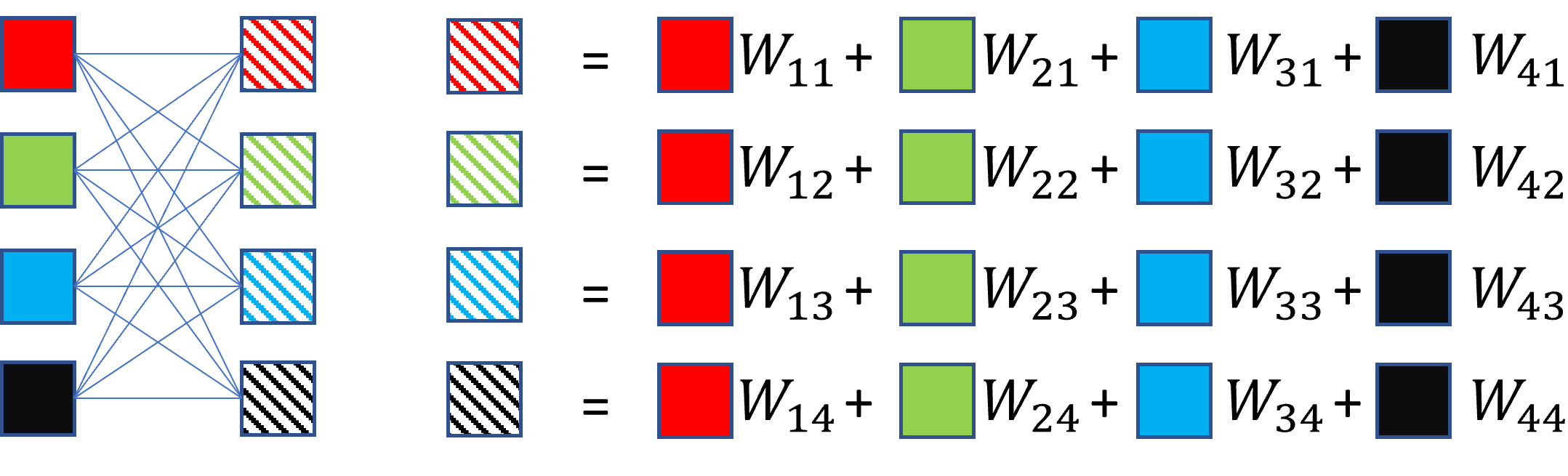} \\
         (a)\\
         \\
         \includegraphics[width=0.45\linewidth]{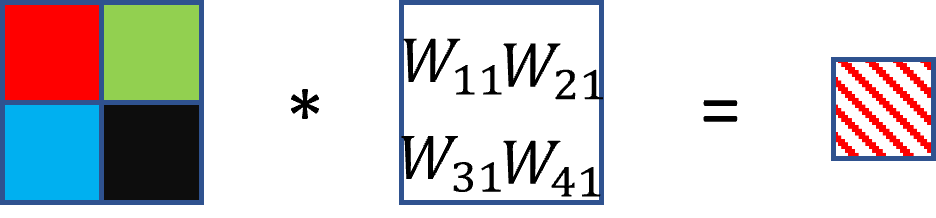} \\
         (b)\\
         \\
         \includegraphics[width=0.9\linewidth]{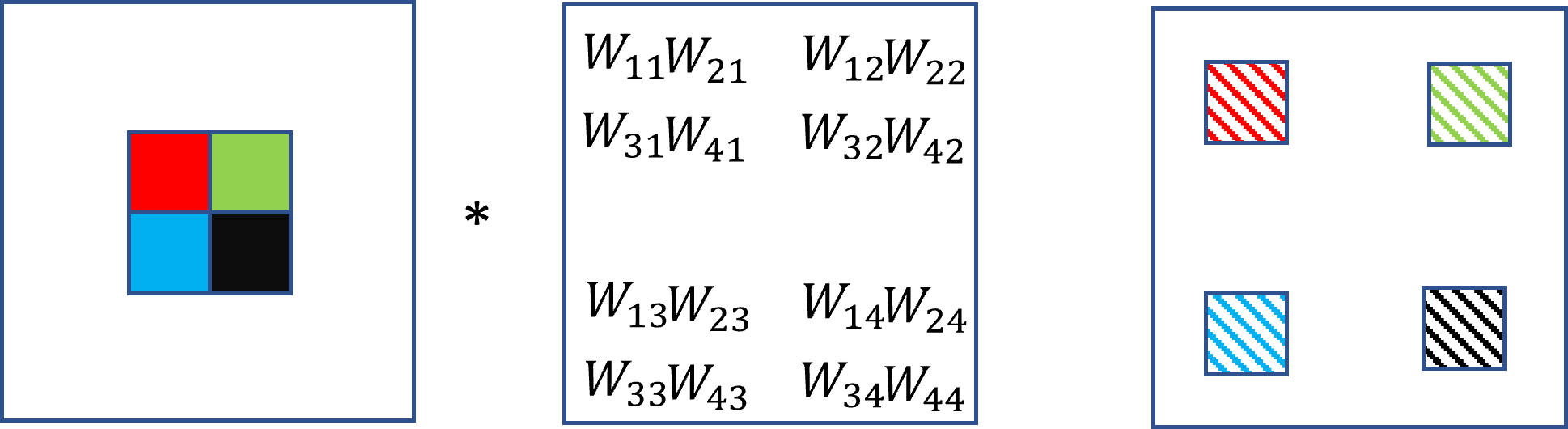} \\
         (c)
    \end{tabular}
    \caption{Implementation of the linear layer using convolution and tiled convolution for 4f system. (a) A simple linear layer of 1 vector. The output vector is the vector-matrix multiplication of the input vector with the weight matrix. Each output node has its own set of weights. (b) Input nodes are in 2D matrix format, convolved with the kernel of equal size and valid padding. The output is similar to one output pixel of the linear layer. (c) Kernel tiling is used to tile all weights of the linear layer in the kernel block. The input is padded to the required resolution. The output archives all output nodes of the linear layer, with the requirement of reshaping (removes zeros in invalid regions).}
    \label{fig:mlp_emulation_in_optics}
\end{figure}

It is important to note initially that the linear layer is the main component of all layers in the transformer's encoder, as it can be seen from Fig.~\ref{fig:Compparison_Vit_ConvShareViT} and as part of the MHSA from Fig.~\ref{fig:normal_mhsa_diagram}. Each output node of the linear layer is the weighted sum of the input nodes as illustrated in Fig.~\ref{fig:mlp_emulation_in_optics} (a). Since the ConvShareVit model deals with patches as matrices rather than vectors, each patch can be convolved with the weight matrix of the same resolution, with valid padding to achieve the output node, as shown in Fig.~\ref{fig:mlp_emulation_in_optics} (b). If the same padding is used, the middle pixel of the output will be the valid region. As shown in Fig.~\ref{fig:mlp_emulation_in_optics} (c), when applied optically, kernel tiling allows for the simultaneous computation of all output pixels.

\begin{figure}
    \centering
    \begin{tabular}{l}
         \includegraphics[width=0.70\linewidth]{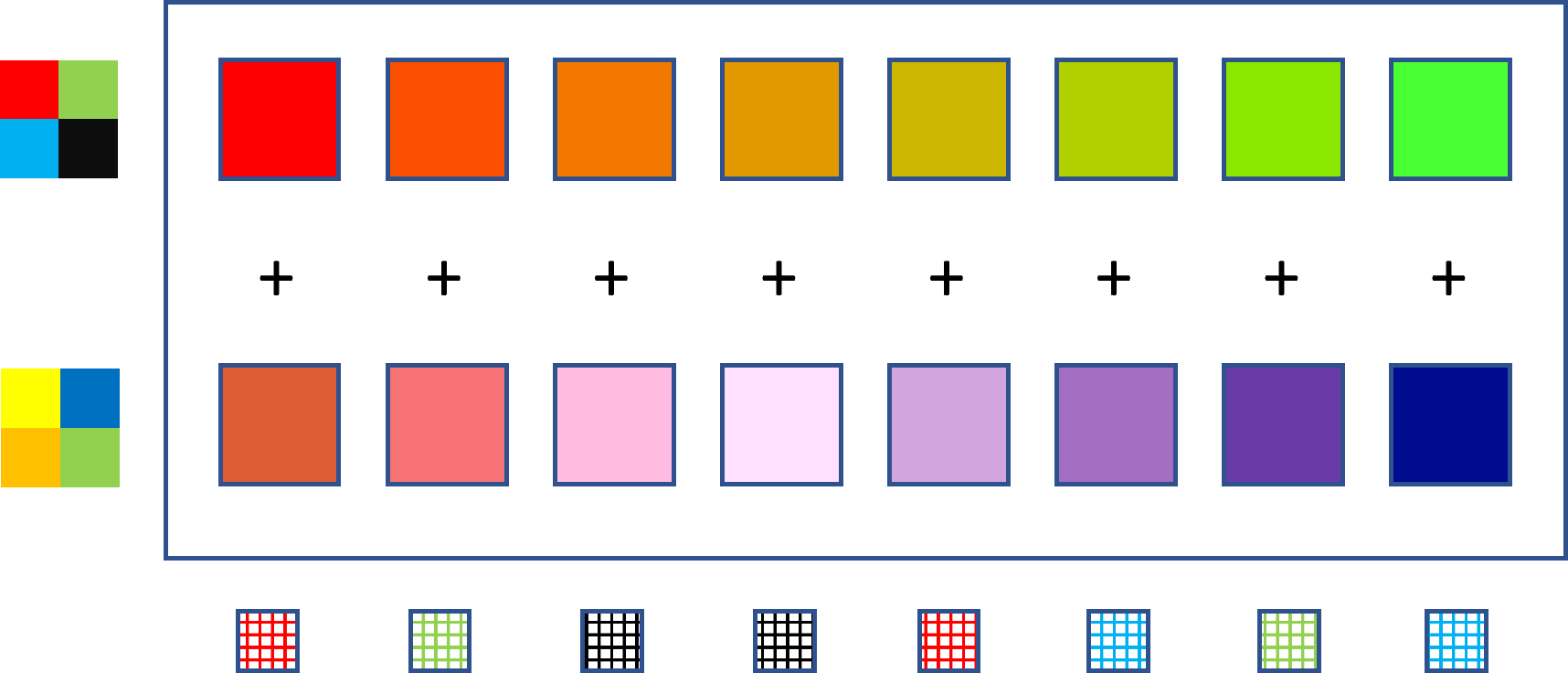} \\
         (a)\\
         \\
         \includegraphics[width=0.70\linewidth]{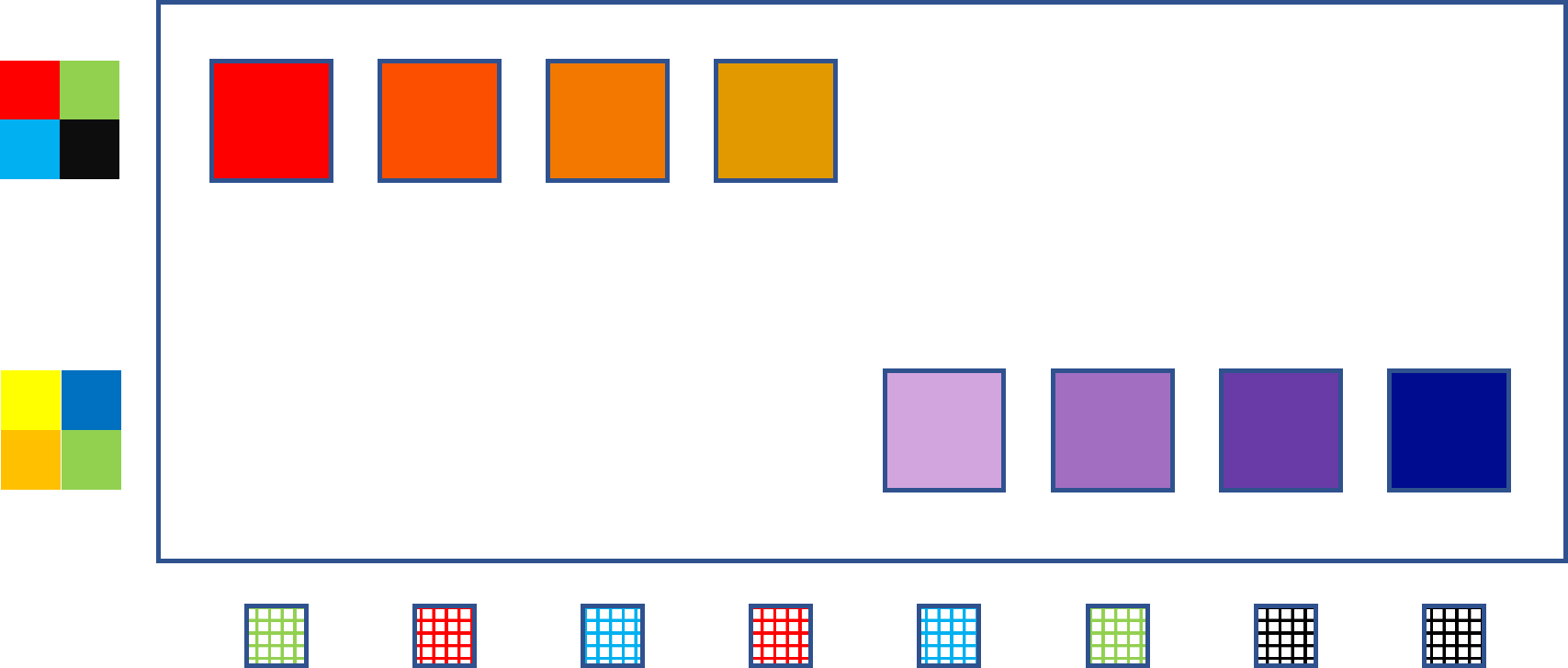} \\
         (b)\\
         \\
         \includegraphics[width=0.94\linewidth]{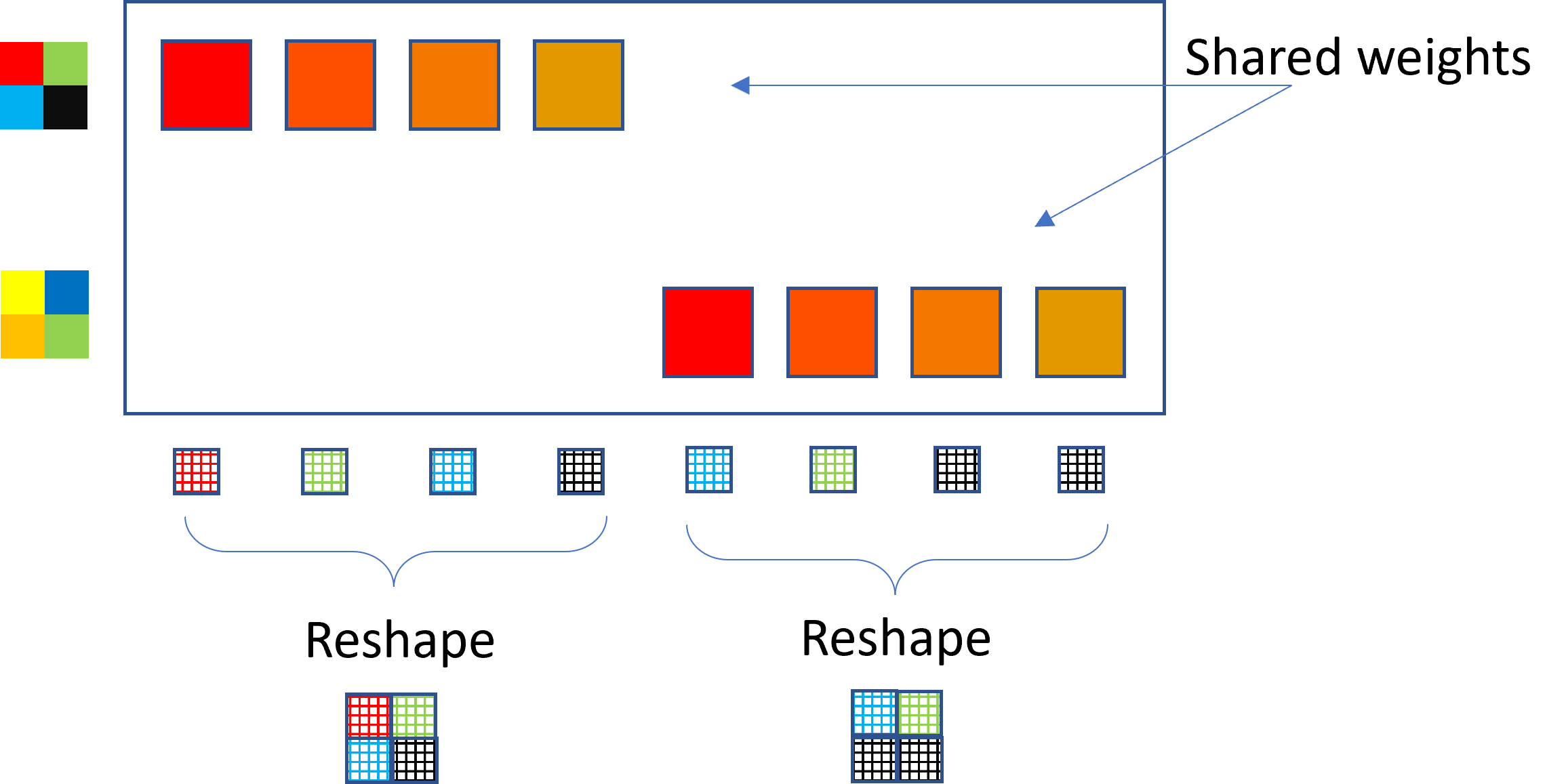} \\
         (c)
    \end{tabular}
    \caption{Shared depthwise convolutional layer, copies the weights across all input channels. (a) Regular convolutional layer, with the groups=1. The number of 2D kernels is equal to the number of input channels $\times$ the number of output channels. (b) Depthwise convolution, where the number of groups is equal to the number of input channels. In this case, each output channel gets only one 2D kernel, meaning no channel summation happens. (c) In the shared depthwise convolutional layer, unlike the regular depthwise convolutional layer, the weights are shared across input channels, making it ideal for the emulation of the Linear Layer. If the kernels are the same resolution as inputs, the valid convolution yields one pixel for each output channel, which can be reshaped into the initial resolution.}
    \label{fig:how_shared_dw_conv_developed}
\end{figure}

In a typical convolutional layer, the kernels are 3D, with the number of kernels matching the number of output channels and the depth of each kernel equal to the number of input channels. Alternatively, this can be viewed as having a set of 2D kernels, where each pair of input and output channels has its own 2D kernel. The results of these 2D convolutions are then summed across the input channels to produce the final output, as shown in Fig.~\ref{fig:how_shared_dw_conv_developed} (a).

In our case, each input channel corresponds to a separate patch that needs to be processed independently, without any interaction between patches. Therefore, summing across channels should be avoided. To achieve this, we use depthwise convolution, where the number of convolution groups is equal to the number of input channels (See Fig.~\ref{fig:how_shared_dw_conv_developed} (b).

However, when a tensor passes through a linear layer, the last dimensions are all mapped into new vectors, meaning the same layer is applied to all dimensions, or in other words, the weights are shared across input channels. Thus, if we emulate a linear layer using convolution, the kernels must be repeated for each input channel during depthwise convolution. This approach, shown in Fig.~\ref{fig:how_shared_dw_conv_developed} (c), is what we refer to as shared depthwise convolution.

\subsection{Multi-head self attention}

\begin{figure}[h]
    \centering
    \begin{tabular}{c}
        \includegraphics[width=0.95\linewidth]{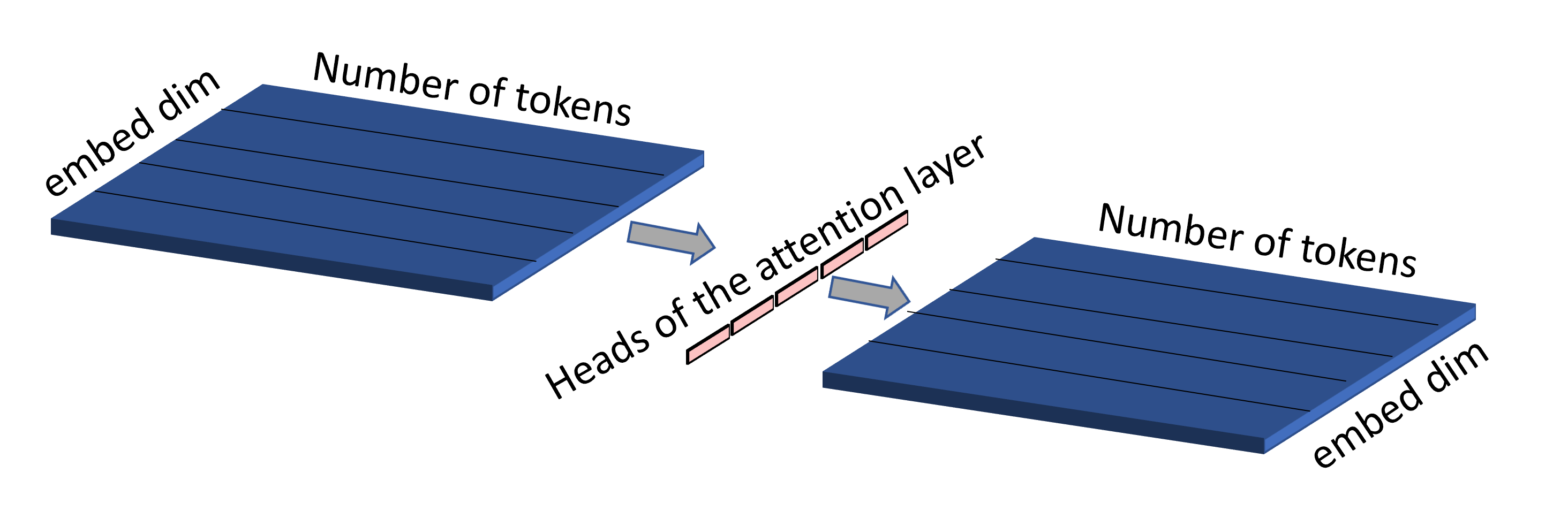}\\
        (a)\\
         \includegraphics[width=0.95\linewidth]{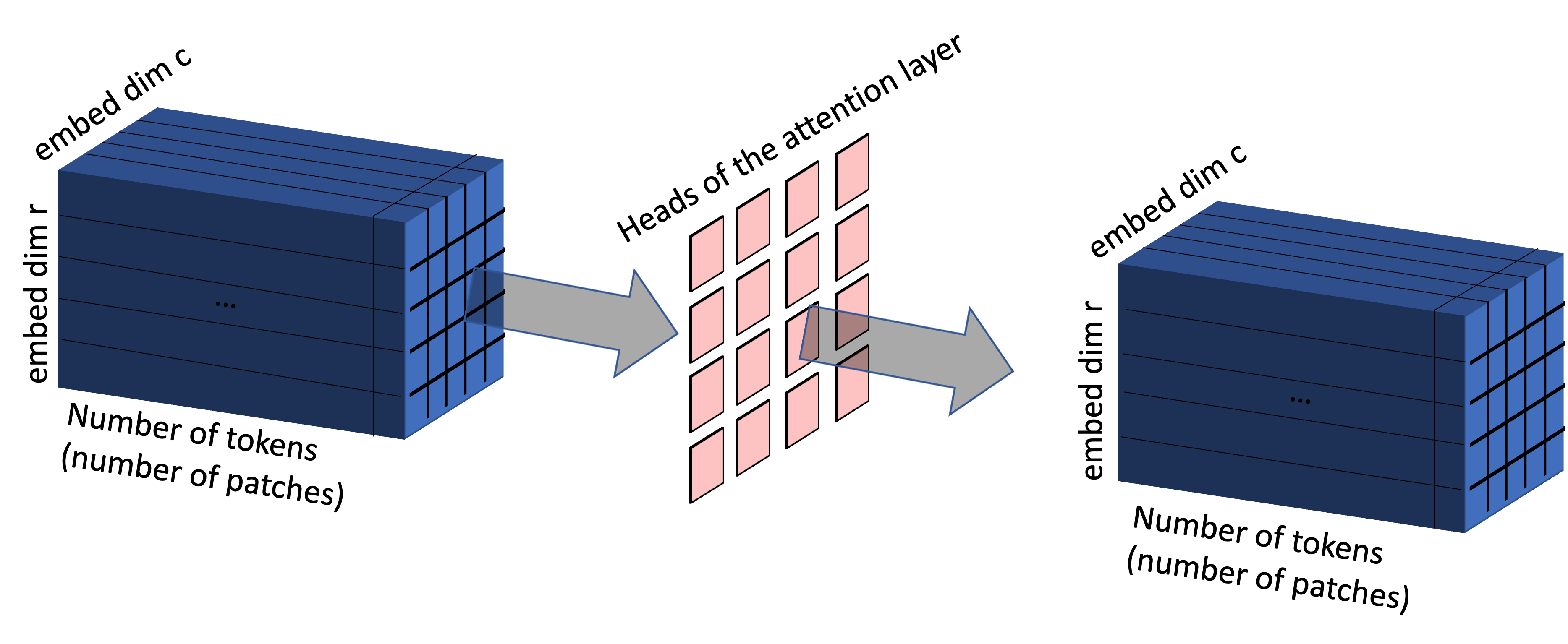} \\
         (b)
    \end{tabular}
    
    \caption{{Visual comparison of input split in regular multi-head attention and our method when the inputs are two-dimensional.} (a) In regular multi-head attention, the input vectors are split into equal-sized vectors, each assigned to a dedicated head of attention, followed by the concatenation of the outputs. (b) In our method, the process can be viewed as patchification, where the two-dimensional input is divided into smaller patches that fit into the heads of convolutional attention layers. The outputs are then merged back into their corresponding locations.}
    \label{fig:multihead_mechanism}
\end{figure}

As it was mentioned before, unlike Vision Transformers, ConvShareViT uses a tensor instead of a matrix as the input. In Vision Transformers, the input is the matrix, where each row vector is the embedding of the input patch. In ConvShareViT, the embedding is represented as the 2D matrix of a 3D tensor, where the depth of the tensor corresponds to the number of patches (tokens). When passed to the different heads of the multi-head self-attention layer, the matrix is split into equal patches, and each patch is fit into its corresponding head. The output of each head is then located at the correct location, where the patch was initially taken from (See Fig.~\ref{fig:multihead_mechanism}). 

% 17.6cm
\begin{figure}
    \centering
    \includegraphics[width=8cm]{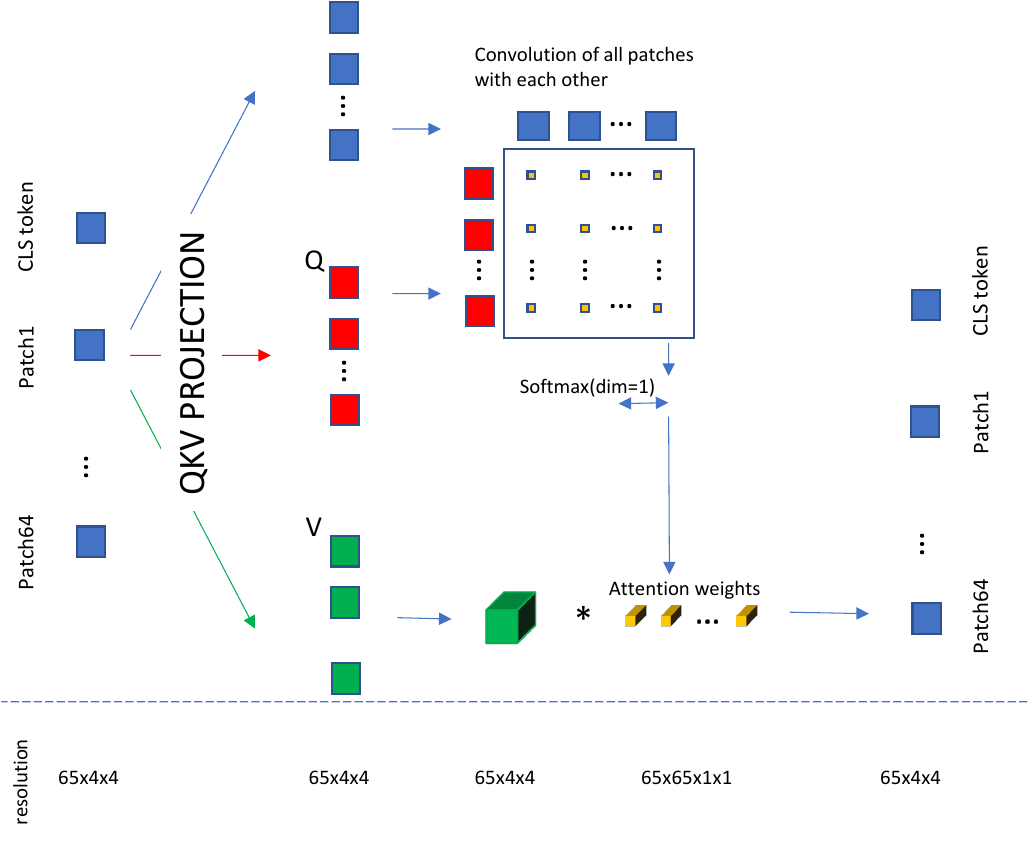}
    \caption{Multi-head self-attention layer using only convolutional layers. The patches of the original tokens are passed through the three shared depthwise convolutional layers, which are reshaped into the original format. The resulting Query and Key matrices then convolved with each other to produce the attention score matrix, which is then softmaxed and used as the weights of the convolutional layer for the values tensor.}
    \label{fig:enter-label}
\end{figure}

When the patches are inputted into the head of the self-attention layer, they go through the $QKV$ projection. In the original Transformer, this stage was performed by passing the inputs through the linear layer. 

This can be replicated by using our shared depthwise convolutional layer, with the kernel resolution equal to the input resolution and the number of kernels (output channels) equal to the number of nodes required at the output of the mimicked linear layer. The spatial resolution of the output of the shared depthwise convolution will be $1 \times 1$ with a number of channels equal to the total number of nodes of the mimicked linear layer, which can be reshaped into the regular matrices. The input tensor is passed through three distinct shared depthwise convolutions in order to achieve Query, Key and Value tensors. 

The next stage is the attention score calculation. In the original Vision Transformer, this is achieved by matrix multiplication of the $Q$ with the transposed $K$ matrix. In other words, the dot product of all vectors $Q$ and $K$. Since in our model, $Q$s and $K$s are matrices rather than vectors, and they are the same resolution, their dot product can be achieved using valid convolution of all to all. The resulting matrix can be scaled, as in the original Vision Transformer, by the square root of the token dimension – in our case, simply the width or height of the token matrix. Following that, the softmax function is applied to the first dimension of the attention score matrix, similar to the regular MHSA layer. 

The last step is the weighted sum of the attention scores, which we do using a point-wise convolutional layer, treating the attention scores as weights of the layer. The outputs are then located in the correct location of the main larger patches. 

In the regular self-attention layer, this last step is performed by a simple matrix multiplication: $Y = A \times V,$
%\begin{equation}
%    Y = A \times V,
%\end{equation}
where: $A$ is the attention score tensor, $V$ is the tensor of values. To do this using the convolution, it first needs to look at the  general formula for the 2D convolutional layer without  bias: 

\begin{equation}
Y_{k, p, q} = \sum_{c=1}^{C_{in}} \sum_{i=0}^{H-1} \sum_{j=0}^{W-1} X_{c, p+i, q+j} \times W_{k, c, i, j},
\end{equation}
where, $X_{c,p+i,q+j}$ refers to the input feature map with the dimension Channels, Height, Width ( $[C_{in},H_{in},W_{in}]$), and the output $Y$ with the dimensions $[C_{out},H_{out},W_{out}]$. W represents the set of kernels with the dimension $[C_{out}, C_{in},H,W]$.

When the convolution is 1D, the spatial dimension is reduced to one, the convolution is simplified to:

\begin{equation}
Y_{k, p} = \sum_{c=1}^{C_{in}} \sum_{i=-a}^{a} X_{c, p+i} \times W_{k, c, i}\text{.}
\end{equation}

When the convolutional layer uses 1x1 kernels, the convolution effectively becomes a point-wise matrix multiplication across channels, identical to dense layer operations. For 1D convolution, the formula with 1x1 kernels becomes:

\begin{equation}
Y_{k, p} = \sum_{c=1}^{C_{in}} X_{c, p} \times W_{k, c},
\end{equation}

which is equivalent to $W \times X$. This leads to the conclusion that matrix multiplication can be treated as the convolutional layer, with the left term being a weight matrix.

\subsection{Multi Layer Perceptron}

In a vanilla transformer encoder, each multi-head self-attention layer is followed by an MLP layer. This MLP usually consists of two linear layers; the first one maps the embedding vectors into the high-dimensional space, and the second one maps it back to the original dimension. One of the hyperparameters of the MLP is the MLP ratio, which indicates how many times the dimension is scaled, meaning the ratio of the hidden layer to the input or output layer. Since the original transformer uses a linear layer, our method leverages the same concept used for the $QKV$ projection in the MHSA layer. 

First, a depthwise convolutional layer with a kernel size equal to the input size is used to map the input into the higher dimension. The output of this layer's resolution is $1 \times 1$ and the number of channels equals to $ (\text{number of tokens} * \text{mlp\_ratio} * H * W)$. This is then reshaped into the $H \times W$ with the channels number of $( \text{number of tokens} * \text{mlp\_ratio})$, increasing the number of tokens by a factor of MLP ratio. 

Similarly, the output resolution of the second shared depthwise convolutional layer is $1 \times 1$ and number of channels $(\text{number of tokens} * H * W)$, which can be reshaped to the input's original shape  \(H \times W \times \text{number of tokens}\).

\begin{figure}[h!]
    \centering
    \begin{tabular}{c}
         \includegraphics[width=8.2cm]{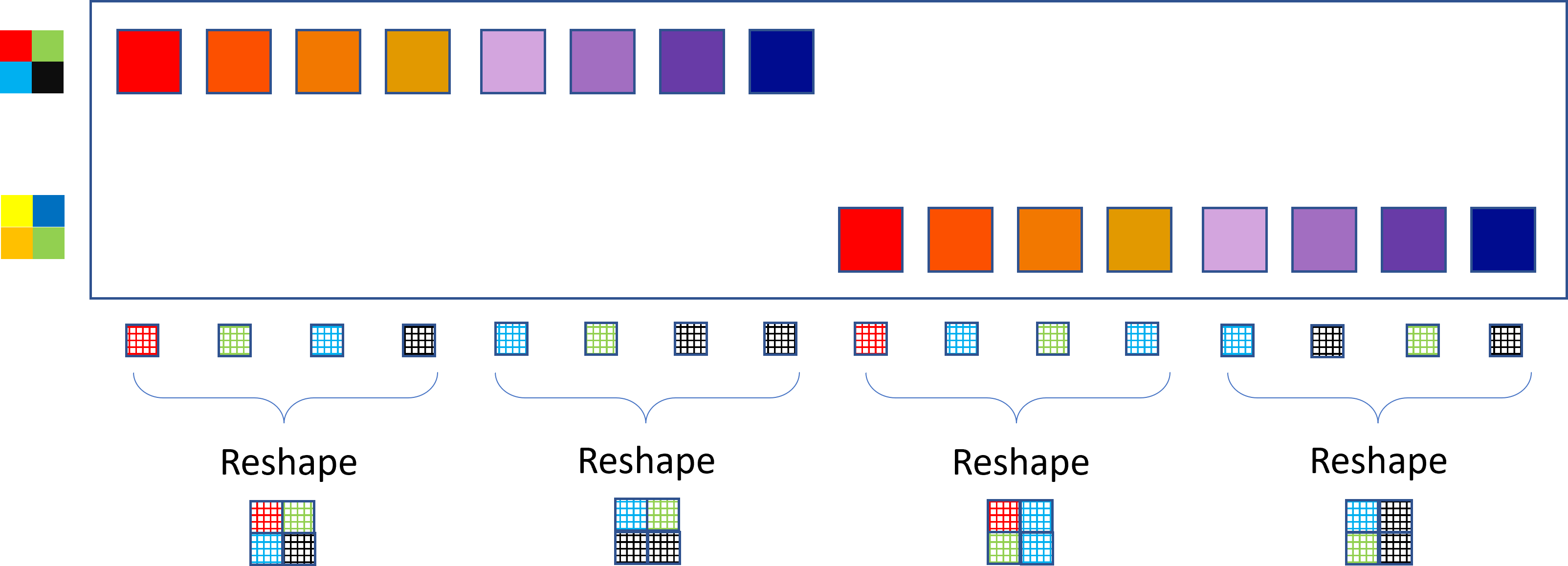}\\
         (a) \\ \\
         \includegraphics[width=4.95cm]{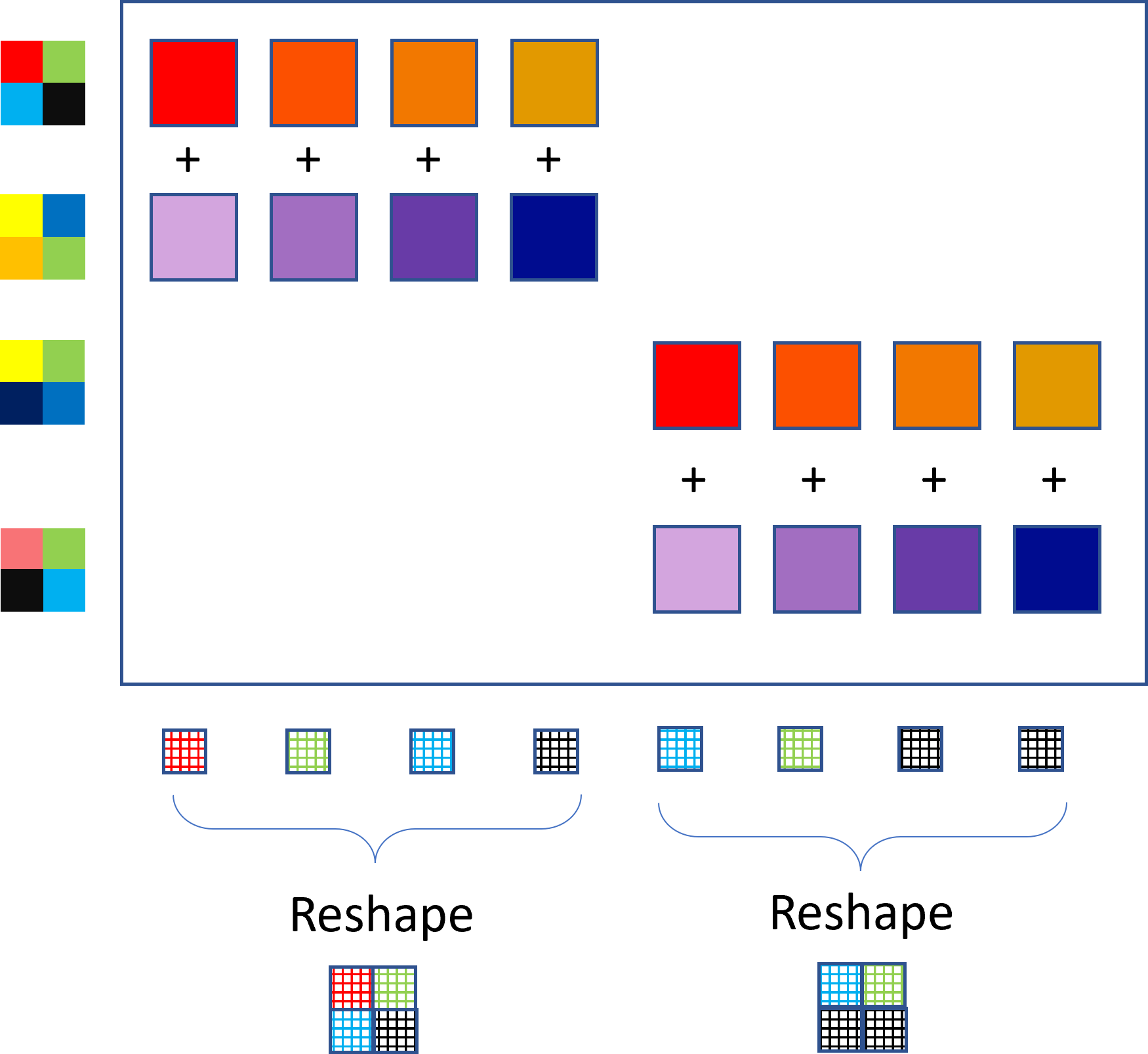} \\
         (b) 
    \end{tabular}
    \caption{{Shared depthwise convolutional layer with the valid convolution and reshape of the output for full emulation of the Linear layer using convolution.} (a) Shared depthwise convolutional layer from one matrix into two matrices. In this case, two matrices have been mapped into 4, where each has been mapped into corresponding two outputs. The technique can be used from few to many mapping. (b) Shared depthwise convolutional layer from two matrices into one. In this case, four matrices have been mapped into two, each group of two into one corresponding output matrix. The technique can be used from many to fewer matrices.}
    \label{fig:shared_depthwise_conv2}
\end{figure}

\subsection{Theoretical parallelism in the 4f system}

The motivation for integrating convolution operations within attention layers is not only to explore the possibility of integrating convolutions into the MHSA but also to exploit the 4f system's capability to perform these tasks more rapidly and efficiently than conventional electronic components. A key benefit of free-space optics is its ability to carry out high-resolution operations without incurring latency. 
% #todo: say how others achieved pararelism

In contrast to CNNs, which rely on convolutional layers, Transformers use linear layers and are typically more efficient on GPUs, where entire tensors are loaded for optimised and rapid computation. However, in an optical setup, the standard input tiling method described earlier becomes inadequate. Parallelisation in this context can be achieved using mixed tiling following the approach of Li~\etal~\cite{li_channel_2020}.

\begin{figure}[h!]
    \centering
    \begin{tabular}{cc}
         \includegraphics[width=0.45\linewidth]{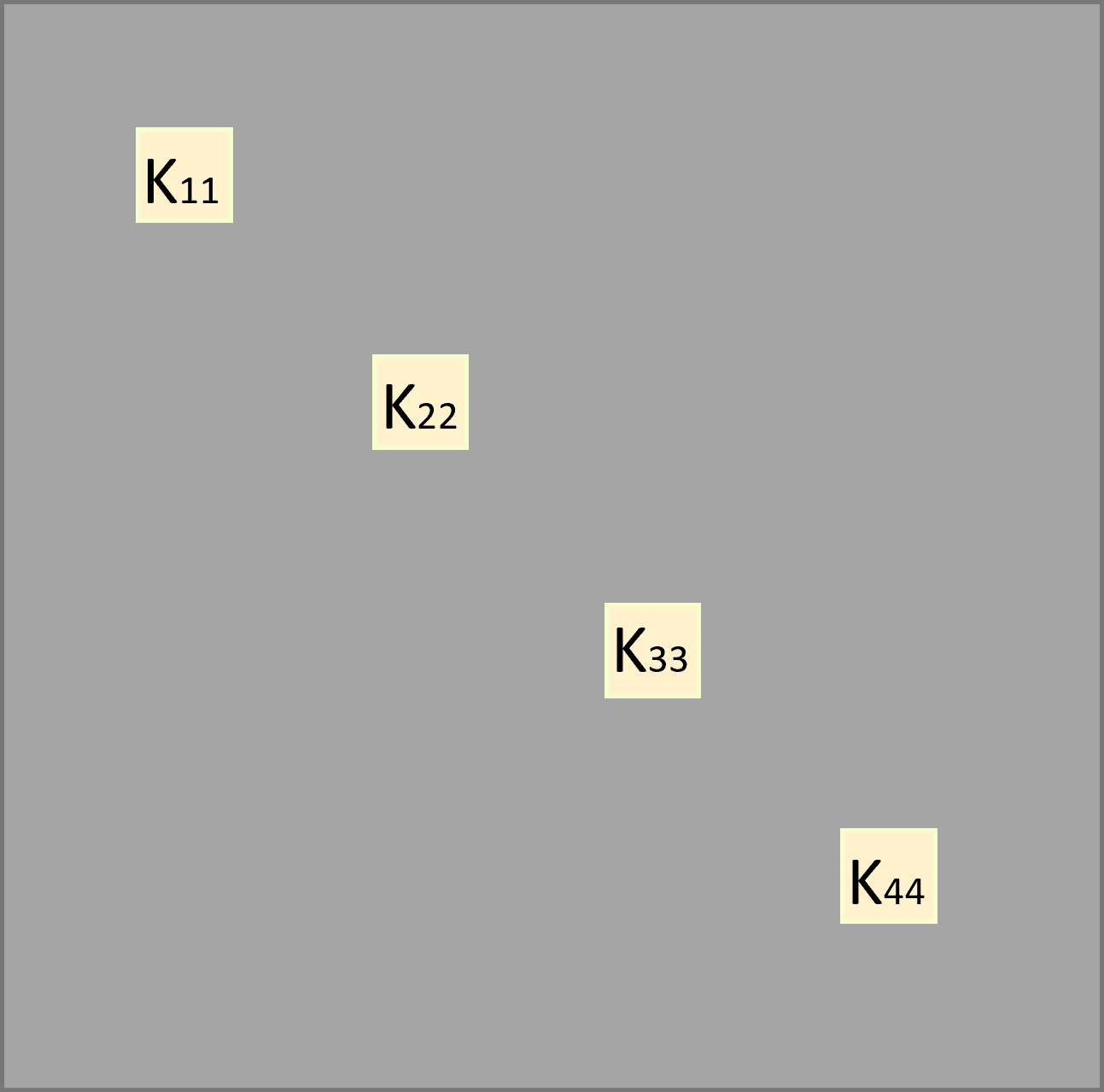}& \includegraphics[width=0.45\linewidth]{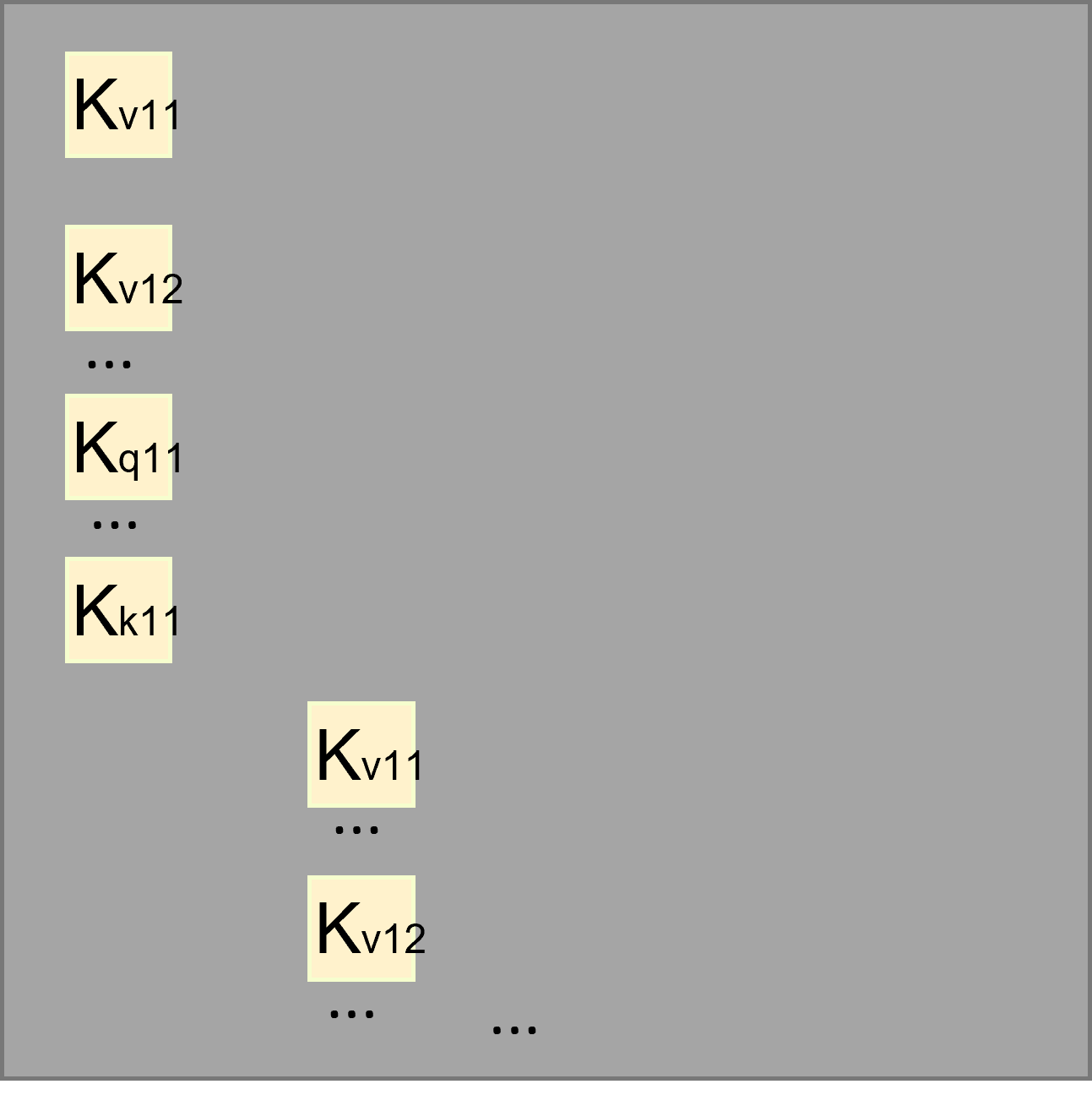} \\
         (a)&(b) 
    \end{tabular}

    \caption{{Mixed tiling with depthwise convolutional layers and its application in $QKV$ projection for convolutional attention layers.} (a) Basic mixed tiling of kernels, with all kernels except those corresponding to the output channel set to zero to prevent summation. (b) Illustration of how shared depthwise convolutions can be applied in the $QKV$ projection.}
    \label{fig:qkv_tiling_mix}
\end{figure}

While mixed tiling is generally used for standard convolution operations, here it must be adapted for depth-wise convolutional layers. This adaptation is accomplished by zeroing out all kernels except the one associated with the output channel, as shown in Fig.~\ref{fig:qkv_tiling_mix} (a).

In $QKV$ projection, the kernels for the $Q$, $K$, and $V$ output pixels can be tiled within a single mixed tiling block and then separated post-output, as shown in Fig.~\ref{fig:qkv_tiling_mix} (b). In most cases, the number of kernels will likely exceed the resolution capacity of the 4f system, necessitating multiple inferences.

% todo in literature review, say how the tiling is done

\section{Experiments}

In this work we performed 4 experiments with ViT and 12 experiments with our method, ConvShareViT.

Initially, four standard ViTs were trained to establish baseline performance, shown in the first part of Table~\ref{tab:models_for_covit}. These included combinations of ViTs with and without trainable positional encoders, as well as models using either multi-head (12 heads) or single-head attention mechanisms. The image patch size was fixed at $4\times4$ from the original image resolution of $32\times32$, resulting in 64 patches (65 when including the classification token). Tokens were embedded into a 192-dimensional space, similar to Lee~\etal~\cite{lee_improving_2022}. Each model comprised 9 transformer blocks with an MLP ratio of 2. Training was conducted for 310 epochs, with the first 10 epochs allocated for warmup~\cite{loshchilov_sgdr_2017}. The Adam optimiser~\cite{Kingma2} was used, starting with a learning rate of $5 \times 10^{-4}$ and using a Cosine Annealing Scheduler~\cite{loshchilov_sgdr_2017}.

For the ConvShareVit models, the main twelve experiments consisting of different methods of application for MHSA, MLP, Shared depthwise convolution and valid convolutions are summarised in the second part of Table~\ref{tab:models_for_covit}. Each model represents a distinct combination of these methods, enabling a systematic analysis of their effects on model performance. Additional details, such as embedding dimensions, the absence of bias and the use of the trainable positional encoder in certain models, are also provided to offer a comprehensive view of the configurations tested. All models were trained using the Adam optimiser, with a learning rate of $5 \times 10^{-4}$~for Models 1–5 and $8 \times 10^{-4}$ for Models 6–12.

\begin{table*}[h!]
    \centering
    \begin{tabular}{ccccccccc}
    \hline
         Model& \makecell{Trainable \\ Pos Encoding}&\makecell{Number of \\ Heads} & MLP used & \makecell{Shared dw\\ Convolution} & \makecell{Valid \\ Convolution} & Bias & Embed Dimension & Accuracy (\%)\\
         \hline \\
            ViT 1& \checkmark& 12&\checkmark&-&-&&192&61\\
            ViT 2& \checkmark& 1 &\checkmark&-&-&&192&63\\
            ViT 3&           &12 &\checkmark&-&-&&192&64\\
            ViT 4&           &1  &\checkmark&-&-&&192&64\\
         \hline
         Model 1& \checkmark&     1      & \checkmark & & &     &$16\times16$ & 49\\ 
         Model 2& \checkmark& 16& \checkmark & & &  \checkmark &$16\times16$&42\\ 
         Model 3& \checkmark& 16& \checkmark & & &  \checkmark &$16\times16$&52\\ 
         Model 4& \checkmark& 16& \checkmark & & &  \checkmark &$16\times16$&52\\ 
         Model 5& \checkmark& 16& \checkmark & & \checkmark&  \checkmark &$16\times16$&48\\ 
         Model 6& \checkmark& 16& \checkmark & &\checkmark & \checkmark & $16\times16$&53\\ 
         Model 7& \checkmark& 16& \checkmark & &\checkmark &  &$16\times16$&54\\ 
         Model 8& \checkmark& 16&  & \checkmark & &  &$16\times16$&25\\ 
         Model 9& \checkmark& 16&  & \checkmark& \checkmark&  &$16\times16$&58\\ 
         Model 10& \checkmark& 1&  & \checkmark& \checkmark&  &$13\times13$&62\\ 
         Model 11& \checkmark& 1&  & \checkmark& \checkmark&  &$16\times16$&63\\
         Model 12&  &1 &  & \checkmark& \checkmark& &$13\times13$ &59\\
        \hline \\
    \end{tabular}
    \caption{Comparison of Methods Applied to Different Models during ConvShareViT development with our implementation of regular ViTs and their performance. This table outlines the primary experiments conducted and the methods applied to each model. Each row represents a distinct model and indicates the presence of specific methods with a checkmark. The differences among Models 2, 3, and 4 lie in the convolutional layers used in the $QKV$ projection.}
    \label{tab:models_for_covit}
\end{table*}

It can be seen that Models 2, 3, and 4 seem identical. The difference lies in the $QKV$ projection, which is listed below:

\begin{itemize}
    \item Model 2 uses a depthwise convolutional layer with same padding and a kernel size equivalent to the patch resolution (4 in this case). The number of input/output channels is the same, corresponding to the number of patches.
    \item Model 3 applies four consecutive depthwise convolutional layers to expand the number of channels by a factor of 7, before reducing it back to the original number and repeats this process.
    \item Model 4 uses two consecutive depthwise convolutional layers to expand the number of channels by a factor of 14 before reducing them back to the original count.
\end{itemize}

A similar pattern is observed with Models 5 and 6, where both models use valid depthwise convolution, reshaping the $1\times1$ outputs to match the original resolution. However, Model 5 includes an additional depthwise convolutional layer, similar to that in Model 2, before the valid convolution is applied

With the ConvShareViTs, the objective was to maintain as much consistency as possible with the original ViTs trained in this study, preserving key aspects such as the MLP ratio, the number of layers, and the patch size of $4\times4$. The primary difference lay in the embedding dimension, as the ConvShareViTs required maintaining a square shape for compatibility, as shown in Fig.~\ref{fig:multihead_mechanism}. Most models, except for Models 10 and 12, had embedding dimensions of 16×16, resulting in a total of 256 dimensions—higher than the original ViTs' embedding dimensions. This increase was essential to enable the extraction of 16 attention heads, each corresponding to a $4\times4$ token. In contrast, Models 10 and 12 utilised an embedding dimension of $13\times13$, which yields 169 dimensions, lower than in our implementations of ViTs. As these models were designed with a single attention head, the tokens did not require further division into patches (Fig.~\ref{fig:multihead_mechanism}(b)), when passed into the convolutional attention layers.

Throughout this study, the CIFAR-100 dataset was employed for all experiments. Data augmentation techniques were applied uniformly across all models, including PyTorch's built-in "CIFAR10" auto-augmentation, random cropping with a padding of 3, random horizontal flipping, and standardisation using the dataset's mean and standard deviation.

\section{Results and Discussion}

\begin{figure}
    \centering
    \begin{tabular}{c}
\includegraphics[width=8.2cm]{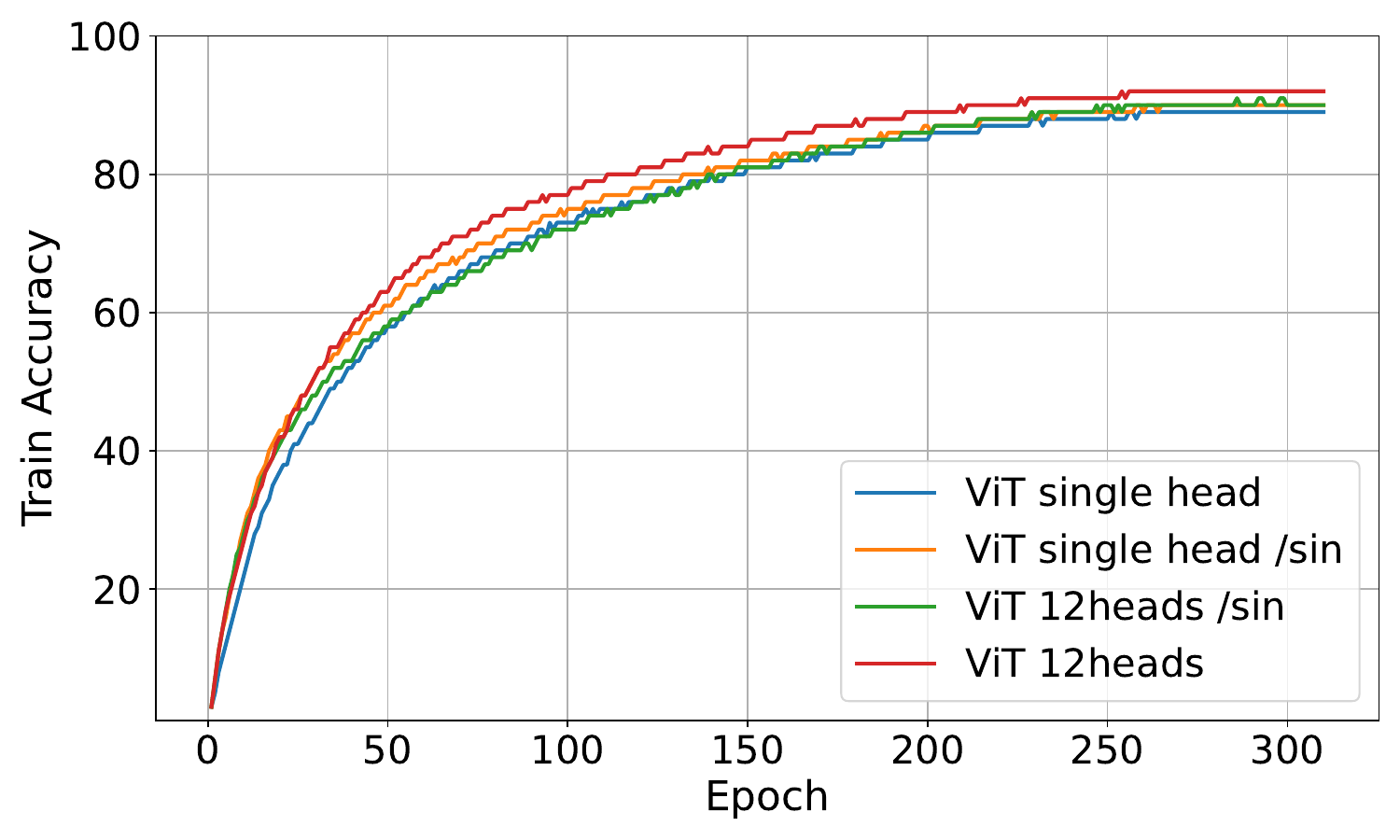} \\
(a)\\ \\\includegraphics[width=8.2cm]{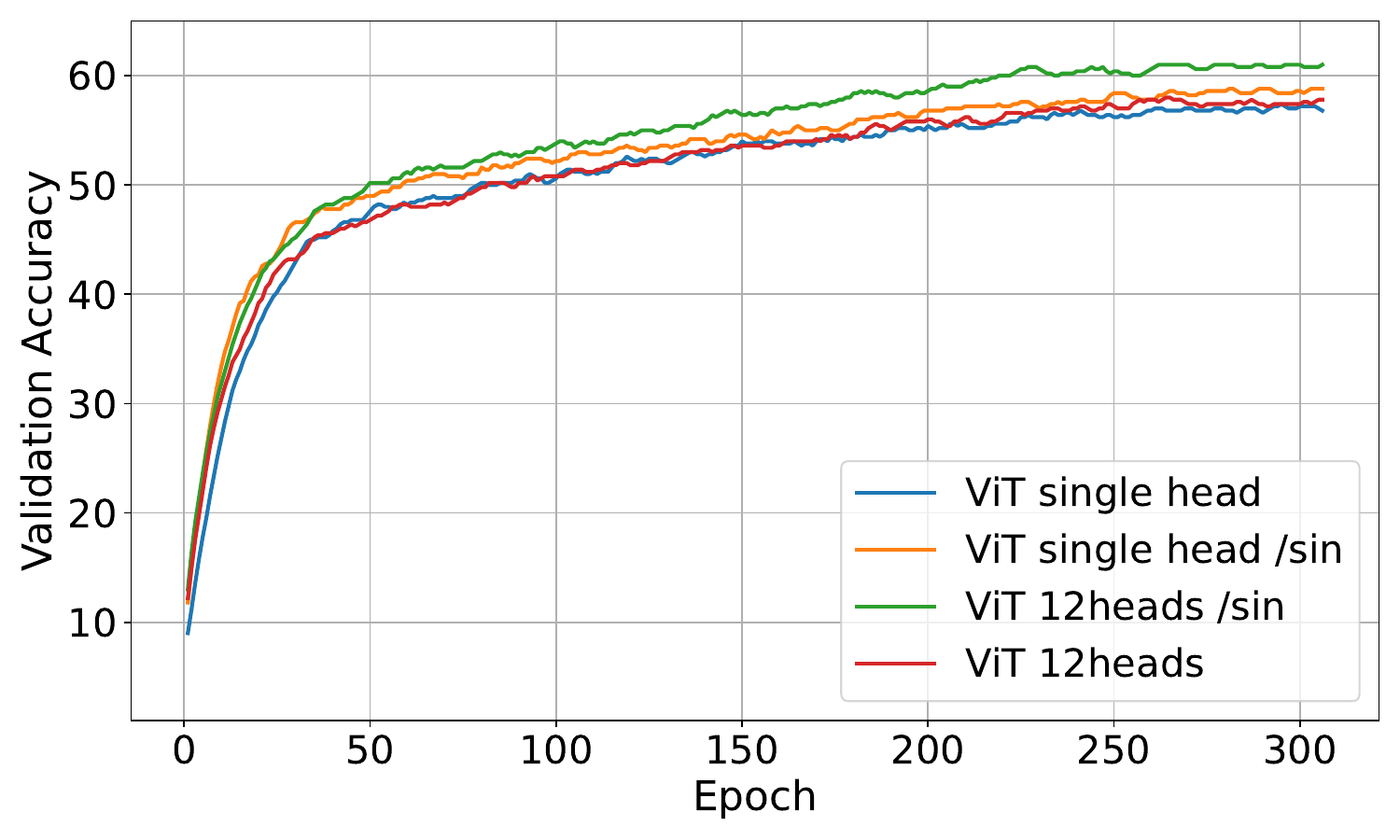} \\(b)
    \end{tabular}
    
    \caption{Comparison of training and validation curves for four ViT models, each using different configurations of positional encoders (trainable vs. sinusoidal / ViT 1, 2 vs ViT 3, 4) and attention heads (single head vs. twelve heads/ ViT 2, 4 vs ViT 1, 3).(a) Training accuracy per epoch. (b) Validation accuracy per epoch.}
    \label{fig:Traincurves_vit}
\end{figure}

The test accuracies of four variations of the regular ViT are shown in the last column of Table~\ref{tab:models_for_covit}. It is clear that the models ViT 3 and ViT 4 using the fixed sinusoidal positional encoder outperform those with trainable position encodings, ViT 1 and ViT 2. This observation aligns with previous findings in Transformer research~\cite{chu_conditional_2022}, suggesting that fixed sinusoidal positional encodings provide more positional information. Furthermore, when the positional encoder is trainable, the single-head model slightly surpasses the twelve-headed model, although the difference is minimal. In contrast, with sinusoidal positional encoding, both models perform nearly identically. Although the test accuracy for both is 64\%, the training curves in Fig.~\ref{fig:Traincurves_vit} illustrate that the twelve-headed ViT with sinusoidal positional encoding demonstrates higher validation performance. On the other hand, the twelve-headed ViT with trainable positional encoding, despite achieving better training performance, shows poor validation accuracy, which is reflected in its test accuracy of 61\%. This suggests that the twelve-headed model might have been over-parameterised for a relatively simple dataset like CIFAR-100, especially when combined with a trainable positional encoder, potentially leading to overfitting. 

% \begin{table}[h]
%     \centering
%     \begin{tabular}{l|llll}
%     \
%          ViT Model&  12 heads & 1 head & 12 heads\_sin& 1 head\_sin\\
%          \hline
%          Acc& 61\% & 63\% & 64\%& 64\%  \\
%      \hline\end{tabular}
%     \caption{Test accuracy on CIFAR-100 of a regular vision transformer with different configurations. Models with "\_sin" indicate the use of sinusoidal position encoding instead of trainable encoding.}
%     \label{tab:vit_perforrmance_our}
% \end{table}

These results can be directly compared with those from prior studies. However, it is important to note that our results are not directly comparable to the results of the original ViT achieved by Dosovitskiy~\etal~\cite{dosovitskiy_image_2021}, nor to state-of-the-art benchmarks for CIFAR-100 classification, as our architecture is novel and does not benefit from pre-trained ViT weights. Dosovitskiy~\etal~themselves emphasised that ViTs require pre-training on large-scale datasets such as ImageNet to achieve strong performance on smaller datasets like CIFAR-100. In contrast, our model is trained entirely from scratch on CIFAR-100, and is thus best compared with studies that similarly train transformer-based models from scratch on this dataset. The selection of the twelve-headed, nine-layer ViT with $4 \times 4$ patches was inspired by the work of Lee~\etal~\cite{lee_improving_2022}. In their study, they also trained ViTs from scratch on CIFAR-100, which is somewhat unusual, as ViTs are typically pre-trained on larger datasets before fine-tuning. Lee~\etal achieved a test accuracy of 60.01\% without augmentation and 73.81\% using several augmentation techniques, including CutMix, Mixup, and AutoAugment. Additionally, they employed methods such as label smoothing, stochastic depth, and random erasing.

In contrast, our approach only used AutoAugment, yet it outperformed Lee's model without augmentation. Although our results are lower than their fully-augmented model, our method still proves effective. The exclusion of CutMix and Mixup was a deliberate choice to maintain training efficiency, as multiple experiments were required to validate our approach. Another study by Zhu~\etal~\cite{zhu_understanding_2023} trained a smaller ViT on CIFAR-100 with six layers and eight heads, achieving a test accuracy of 54.31\%.

Our novel method, ConvShareViT, also yielded specific performance characteristics. Notably, Models 10, 11, and 12, which all employed a single attention head, outperformed the multi-headed models, as shown in Table~\ref{tab:models_for_covit}. Model 8, the only model to use the same padding in the $QKV$ projection, achieved a poor test accuracy of 25\%. This suggests that using valid convolution and reshaping the outputs (effectively mimicking an MLP) is crucial for performance. Interestingly, models up to model 4 also used the same padding but retained the original output shapes, as in Fig.~\ref{fig:how_shared_dw_conv_developed} (b). These models had more trainable parameters due to not sharing weights across input channels, which may explain their relatively high performance. Despite their strong results, these models did not employ traditional attention mechanisms, as revealed by later visualisation analyses.

% \begin{table}[]
%     \centering
%     \begin{tabular}{l|l}

%          Model & Accuracy (\%) \\ 
%          \hline
%          1  & 49  \\
%          2  & 42  \\
%          3  & 52  \\
%          4  & 52  \\
%          5  & 48  \\
%          6  & 53  \\
%          7  & 54  \\
%          8  & 25  \\
%          9  & 58  \\
%          10 & 62  \\
%          11 & \textbf{63} \\
%          12 & 59  \\
%     \end{tabular}
%         \caption{Test Accuracy of models described previously in Table~\ref{tab:models_for_covit}.}

%     \label{tab:my_label}
% \end{table}

\begin{figure}
    \centering
    \begin{tabular}{c}
         \includegraphics[width=0.95\linewidth]{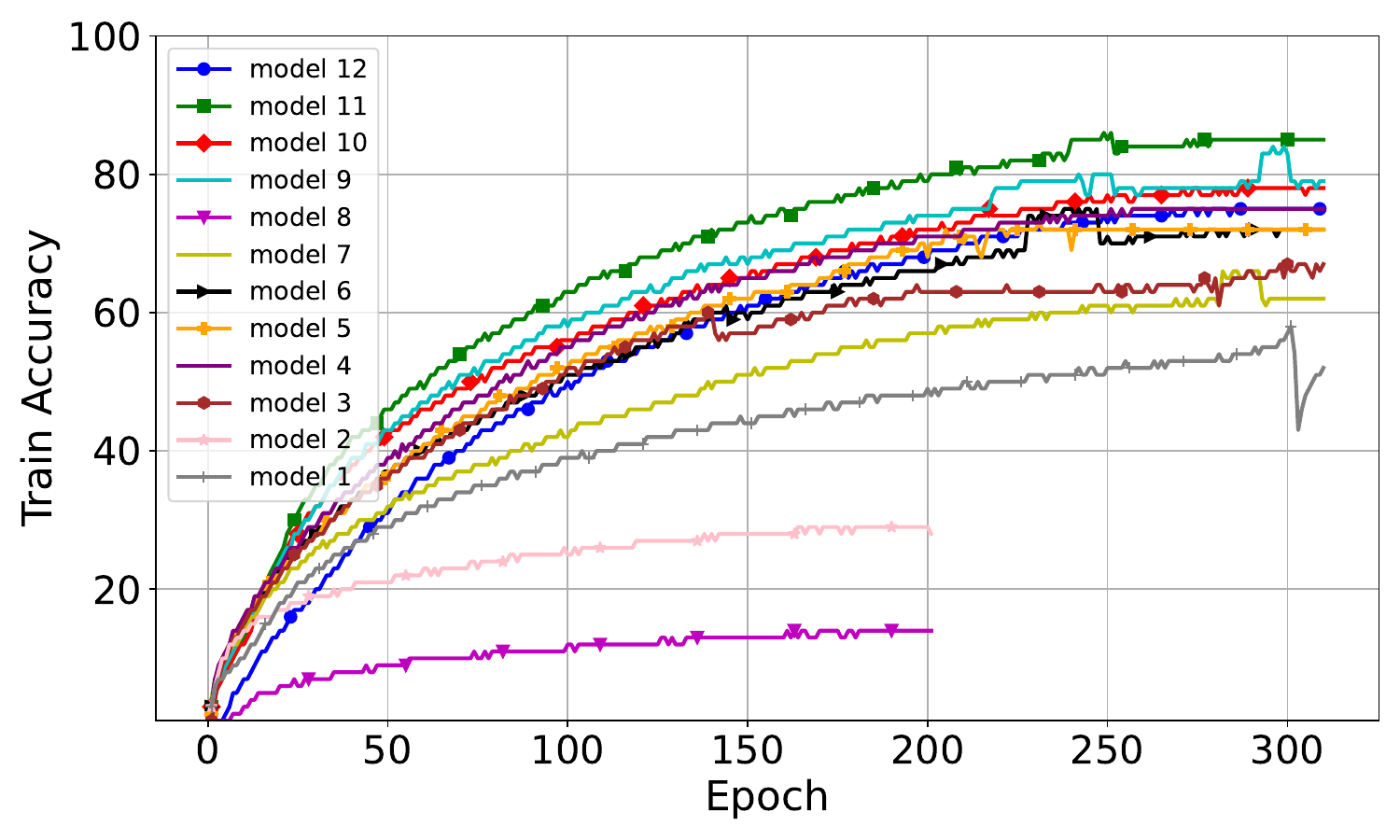}\\
         (a)\\
         ~\includegraphics[width=0.93\linewidth]{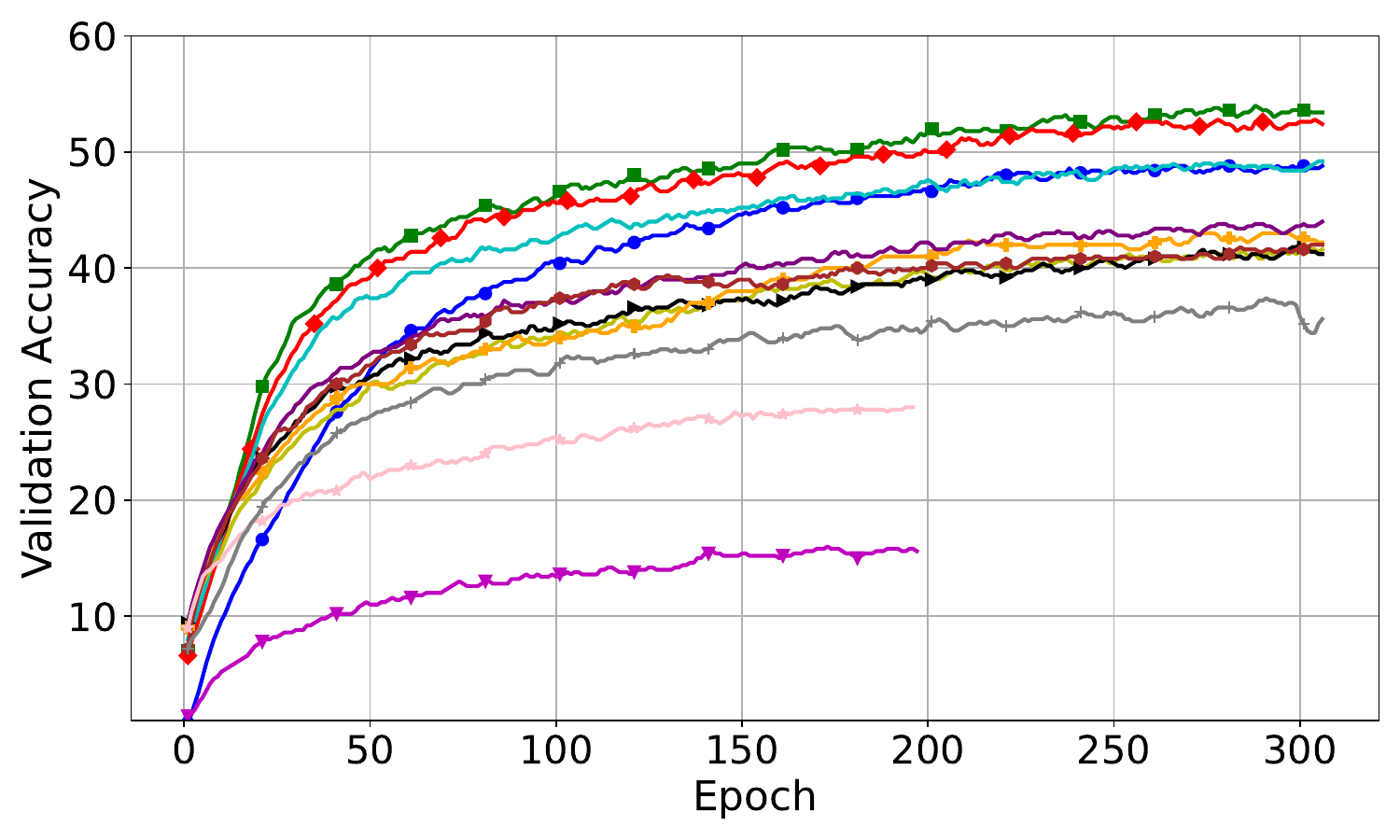}\\
         (b)
    \end{tabular}

    \caption{{Training curves comparison for the Training set and Validation set of CIFAR-100 with different ConvShareVit models} (a) Training curve for train set of CIFAR-100, with the best model being model 11. (b) The training curve for the validation set of CIFAR-100, with the best model being model 11. 
    % It should be noted that the training of Model 2 was terminated early due to its low performance. 
    Models with validation accuracy lower than 30\% after epoch 200 were terminated early due to poor performance.}
    \label{fig:models_training_curve}
\end{figure}

Fig.~\ref{fig:models_training_curve} shows the training process of all models which employed the convolution operation within the MHSA layers. All models were trained for 310 epochs, with 10 epochs reserved for warmup, except for model 2, which was halted early due to poor performance and no improvement in the loss. Model 11 led both in training and validation curves, consistent with its test accuracy in Table~\ref{tab:models_for_covit}. While model 9 appeared second in training accuracy, model 10 outperformed it in validation accuracy. This is also evident from the test accuracies in Table~\ref{tab:models_for_covit}, where models 9 and 10 scored 58\% and 62\%, respectively, suggesting that model 9 may suffer from slight overfitting. It is also worth noting that model 10, which uses a single head and a smaller embedding dimension of 13, performed better than the multi-headed model 9. This reinforces the hypothesis that single-head attention may suffice for classifying CIFAR-100.

\begin{figure*}[!htbp]
    \centering
    \begin{tabular}{ll}
         \begin{turn}{90}ViT\end{turn} & \includegraphics[width=0.95\linewidth]{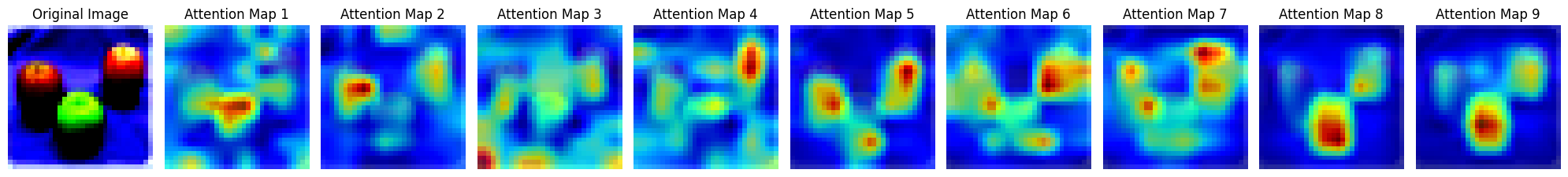} \\
         \begin{turn}{90}M-12\end{turn} & \includegraphics[width=0.95\linewidth]{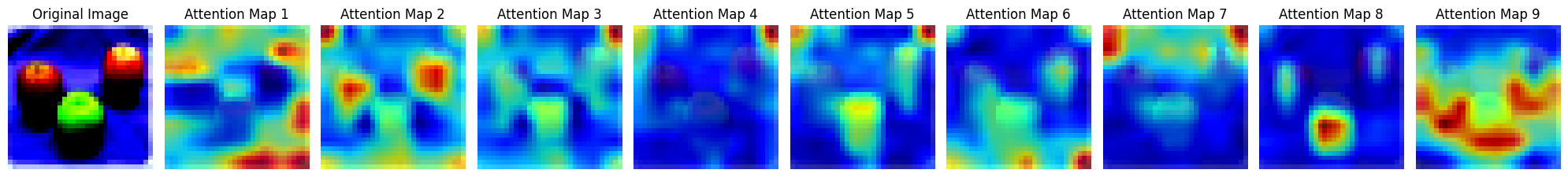}\\
         \begin{turn}{90}M-11\end{turn}& \includegraphics[width=0.95\linewidth]{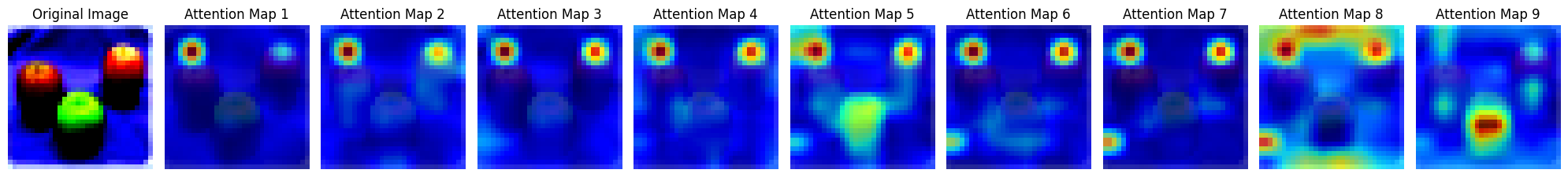}\\
         \begin{turn}{90}M-10\end{turn}& \includegraphics[width=0.95\linewidth]{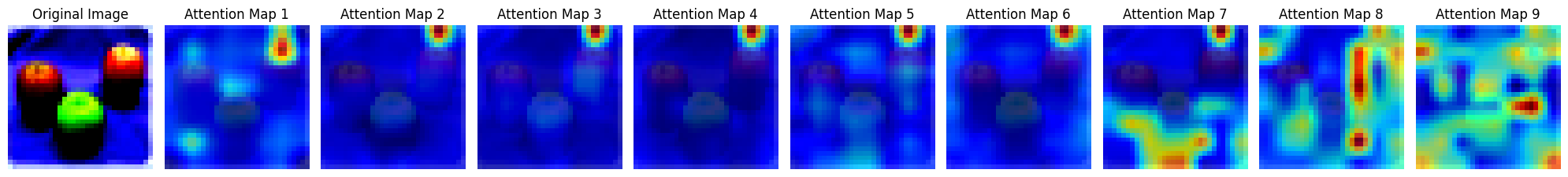} \\
         \begin{turn}{90}M-9\end{turn}& \includegraphics[width=0.95\linewidth]{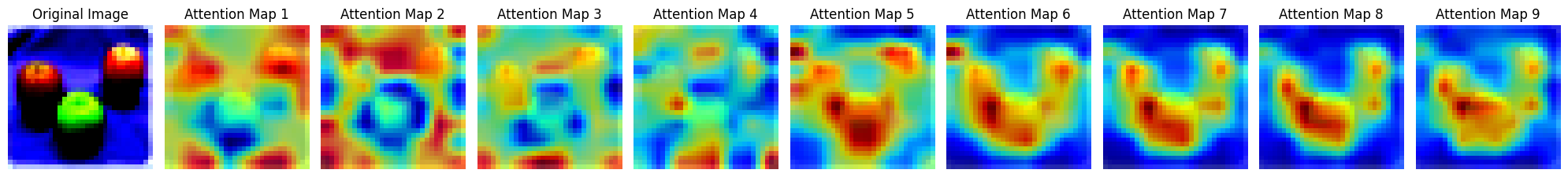}\\
         \begin{turn}{90}M-8\end{turn}& \includegraphics[width=0.95\linewidth]{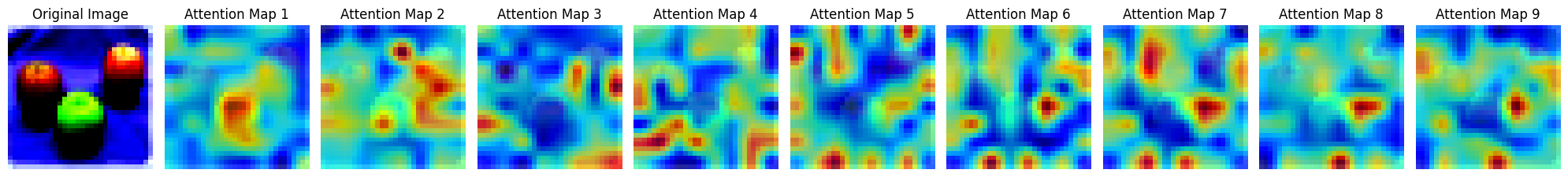}\\
         \begin{turn}{90}M-7\end{turn}& \includegraphics[width=0.95\linewidth]{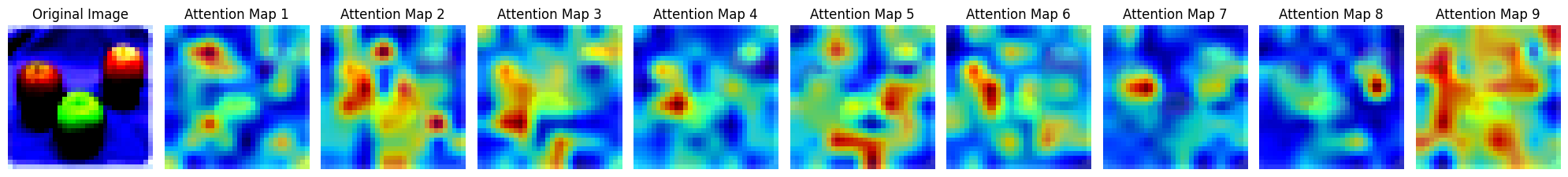}\\
         \begin{turn}{90}M-6\end{turn}& \includegraphics[width=0.95\linewidth]{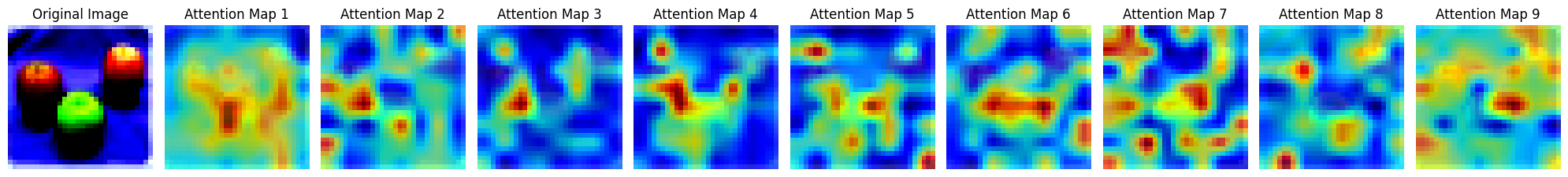}
    \end{tabular}
    
    \caption{{Visualisation of the average attention scores projected onto the original input image of Apples from the test set of CIFAR-100.} This figure compares the performance of the last seven models with the regular ViT (Vision Transformer with 12 heads). The vertical axis corresponds to the models and the horizontal to the attention layers. The ViT model achieved good attention scores in the final layers using a standard attention mechanism. Models 9, 11, and 12 also achieved attention scores similar to the original ViT. In contrast, Model 10's attention scores look incorrect as it is focusing on the background instead, as evidenced by other visualisations. Model 8 did not converge, while Models 7 and 6 did not employ the Shared DW convolutional methods without emulating the linear layer, causing the models to not learn the attention scores in the same manner as the ViT.}
    \label{fig:attnvis_appels}
\end{figure*}

\begin{figure*}[!htbp]
    \centering
    \begin{tabular}{ll}
         \begin{turn}{90}ViT\end{turn} & \includegraphics[width=0.95\linewidth]{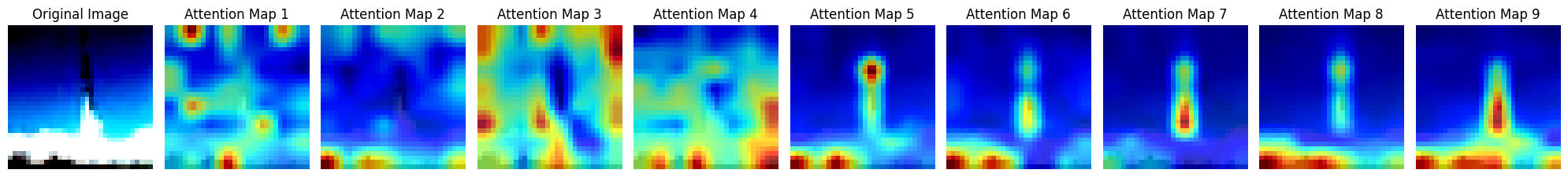} \\
         \begin{turn}{90}M-12\end{turn} & \includegraphics[width=0.95\linewidth]{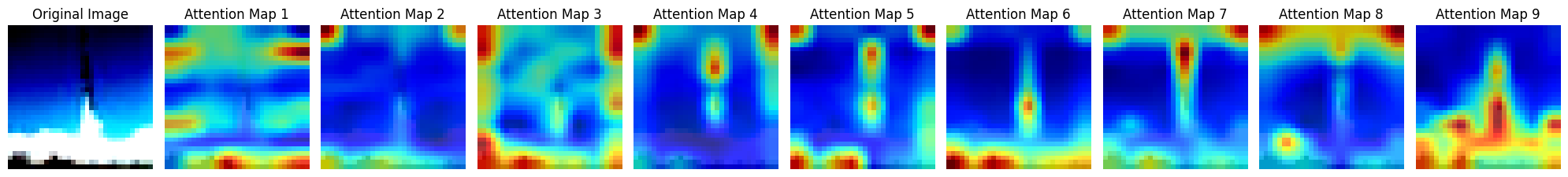}\\
         \begin{turn}{90}M-11\end{turn}& \includegraphics[width=0.95\linewidth]{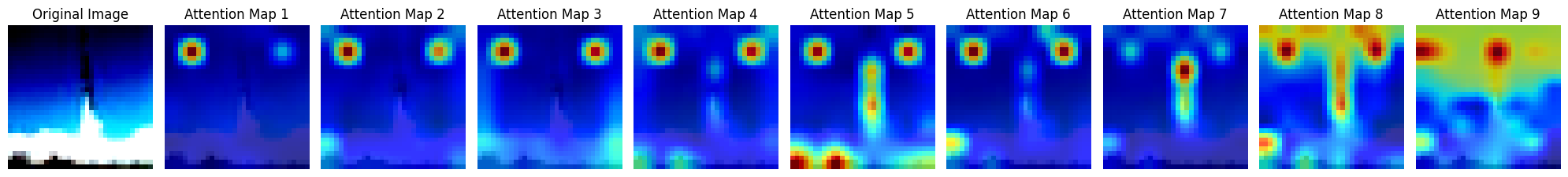}\\
         \begin{turn}{90}M-10\end{turn}& \includegraphics[width=0.95\linewidth]{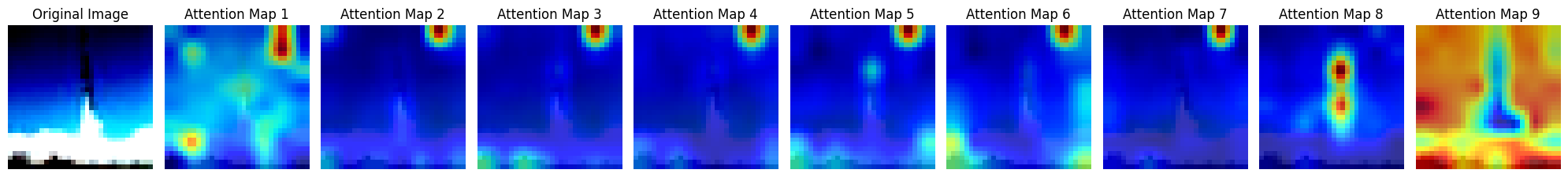} \\
         \begin{turn}{90}M-9\end{turn}& \includegraphics[width=0.95\linewidth]{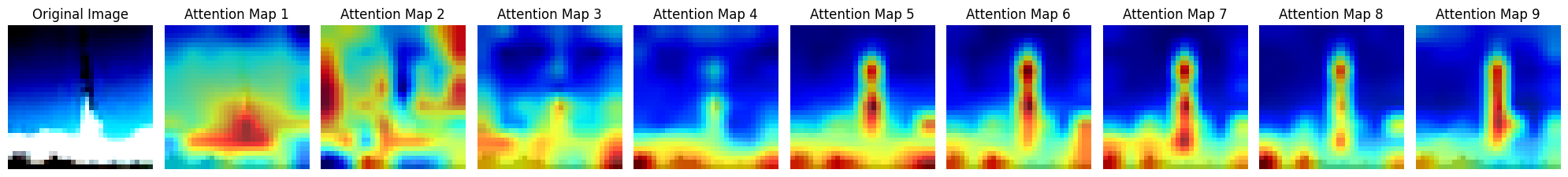}\\
         \begin{turn}{90}M-8\end{turn}& \includegraphics[width=0.95\linewidth]{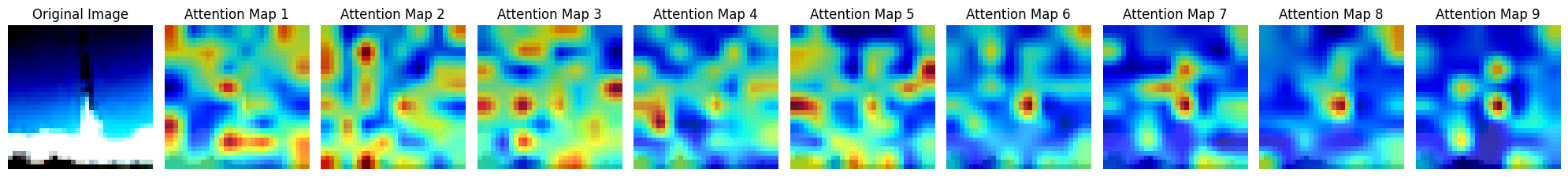}\\
         \begin{turn}{90}M-7\end{turn}& \includegraphics[width=0.95\linewidth]{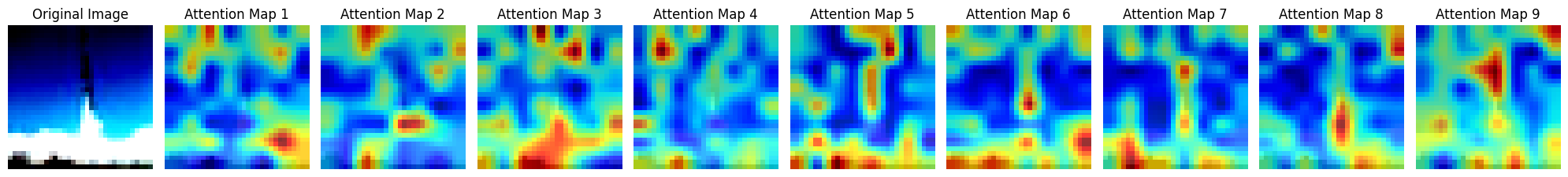}\\
         \begin{turn}{90}M-6\end{turn}& \includegraphics[width=0.95\linewidth]{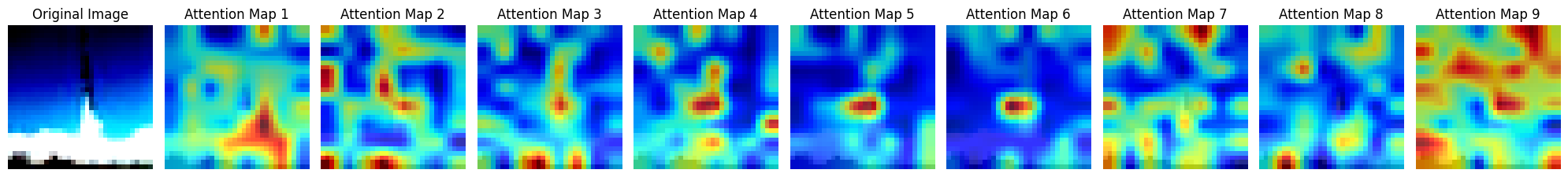}
    \end{tabular}
    
    \caption{{Visualisation of the average attention scores projected onto the original input image of a rocket from the test set of CIFAR-100.} This figure compares the performance of the last seven models with the regular ViT (Vision Transformer with 12 heads). The vertical axis corresponds to the models and the horizontal to the attention layers. The ViT model achieved good attention scores in the final layers using a standard attention mechanism. Models 9, 11, and 12 also achieved attention scores similar to the original ViT. In contrast, Model 10's attention scores focused on the background instead, which still managed to achieve a good performance. Model 8 did not converge, while Models 7 and 6 did not employ the Shared DW convolutional methods without emulating the linear layer, causing the models to not learn the attention scores in the same manner as the ViT.}
    \label{fig:eattnvis_rocket}
\end{figure*}

Figures~\ref{fig:attnvis_appels} and ~\ref{fig:eattnvis_rocket} provide visual comparisons of average attention scores per layer for both the standard ViT and the ConvShareViT models for selected images of "apples" and a "rocket" from CIFAR-100. Notably, only models from Model 8 onwards implement shared depthwise convolution. However, Model 8 does not apply valid convolution with output reshaping, making the visualisation of attention scores only meaningful from Model 9 onwards.

In Model 9, starting with the fifth layer, the attention scores start to concentrate primarily on the objects in the image. It's interesting to note that despite Model 10 performing well, the majority of its attention scores are focused on the background. The top-performing model, model 11, shows good learning, with noticeable attention scores throughout the layers. But Model 11 also demonstrates a bias in its learnable positional encoding, which sustains across all layers and produces high attention scores near the image corners.

To mitigate the issue of biased positional encoding, Model 12 was developed with fixed sinusoidal positional encoding. This change led to more balanced attention scores, though there was a slight drop in test accuracy to 59\%. This performance decline is likely attributable to Model 12's reduced size, as it uses an embedding dimension of $13 \times 13$.

These results suggest that while the models can achieve high accuracy, the use of same padding in shared depthwise convolution yields suboptimal results (Model 8). Models that used same padding but without weight sharing across the input channels performed better than Mode 8, likely because they operate as regular feature extractors as in CNNs. The visualisations indicate that models only effectively learn attention when shared depthwise convolution is combined with valid padding and output reshaping—essentially when convolution functions similarly to a standard linear layer.

This finding implies that the use of convolution in self-attention layers is only beneficial when it can replicate linear layers. In cases where this is not achieved, the convolution behaves more like traditional convolutional neural networks, introducing additional complexity without improving model performance. Although models 7 and earlier converged, it is difficult to categorise them as transformers. Nonetheless, the results confirm that convolution can be used to replicate linear layers by employing shared depthwise convolutional layers, and this method can be effectively integrated into the 4f system.

For efficient parallel inference, as discussed earlier, the most effective setup in a 4f system network is a single-head model with a $13 \times 13$ embedding dimension, similar to models 10 and 12. This setup can be parallelised using mixed tiling. The number of possible input or output channels in mixed tiling depends on the kernel size and input size. The maximum number of input channels or output channels for the convolutions is given by:

\begin{equation}
    n = \left\lfloor\frac{R}{M+N-1}\right\rfloor,
\end{equation}
where $M$ is the input resolution, $N$ is the kernel resolution, $R$ is the device's resolution.

% last two paragprahs not paraphrased yet

In shared depthwise convolution with valid padding, the input size equals the kernel size, which in this case is 13. With an estimated 4K resolution (2160 pixels per dimension), the number of input and output channels is 86. This results in 65 input channels and $65 \times 13^2$ output channels. For the three $QKV$ (Query, Key, Value) projections, 384 inferences are needed. To compute the attention scores, 50 inferences are required, using a 4f system with mixed tiling at 4K resolution. Finally, the weighted sum attention score is computed in one inference using mixed tiled convolution.

For the convolutional MLP blocks, kernel tiling is more efficient. In this case, the required number of input channel inferences is 65 and $65 \times 2$, leading to 195 inferences per block. Across nine layers, this totals 5,670 inferences for the 4f system. On a 2 MHz device~\cite{gupta_4f_2022,li_channel_2020}, this process would take \textbf{2.8 ms}, compared to measured \textbf{8.5 ms} on a T4 GPU, based on 700 inference runs, with the GPU warmed up for 10 iterations beforehand.

\section{Conclusion}
In this work we have presented a ConvShareViT architecuture, which is a Vision Transformer adapted for the 4f free-space optical accelerators to enhance neural network speed by taking advantage of the high-resolution capabilities of free-space optics. ViTs were trained from scratch on the CIFAR-100 dataset, and their performance was compared with twelve models incorporating convolutional operations into the attention mechanism. The only models that effectively learned the attention mechanism were those employing the newly developed shared depthwise convolutional layers with valid padding, reshaped to mimic linear layers, as evidenced by the average attention score visualisations.

The study successfully demonstrated the viability of training ViT models using only convolutional layers within the 4f system for acceleration. The changes in performance and increase in the inference speed were investigated with various tiling methods, including mixed and kernel tiling. These techniques broaden the potential of 4f free-space optical acceleration to transformer-based models like ViTs, moving beyond traditional CNN architectures.

Future work could explore approaches similar to FatNet~\cite{ibadulla_fat-u-net_2024, ibadulla_fatnet_2023} to further optimise ConvShareViTs for minimal inference use by taking advantage of the high-resolution capabilities of the 4f system for additional performance improvements. Moreover, more efficient methods of running the models could be investigated, particularly to evaluate their performance on a simulator or in a real 4f free-space optics environment. Additionally, the effects of noise and misalignment of the elements during optical training could be explored in relation to the models used in this work.

\section*{Acknowledgment}
We would like to dedicate this work to the memory of Professor Thomas M. Chen, whose guidance and mentorship played a pivotal role in shaping this research. His contributions to the fields of cybersecurity and AI were matched only by his generosity as a teacher and colleague. He will be greatly missed by all who knew him.
\bibliographystyle{unsrt}

%{\appendices
%\section*{Proof of the First Zonklar Equation}
%Appendix one text goes here.
% You can choose not to have a title for an appendix if you want by leaving the argument blank
%\section*{Proof of the Second Zonklar Equation}
%Appendix two text goes here.}

% You can use a bibliography generated by BibTeX as a .bbl file.
%  BibTeX documentation can be easily obtained at:
%  http://mirror.ctan.org/biblio/bibtex/contrib/doc/
%  The IEEEtran BibTeX style support page is:
%  http://www.michaelshell.org/tex/ieeetran/bibtex/
 
%  % argument is your BibTeX string definitions and bibliography database(s)
% %\bibliography{IEEEabrv,../bib/paper}
% %
% \section{Simple References}
% You can manually copy in the resultant .bbl file and set second argument of $\backslash${\tt{begin}} to the number of references
%  (used to reserve space for the reference number labels box).

% \bibliographystyle{IEEEtran}
\bibliography{main}

% If you have an EPS/PDF photo (graphicx package needed), extra braces are
%  needed around the contents of the optional argument to biography to prevent
%  the LaTeX parser from getting confused when it sees the complicated
%  $\backslash${\tt{includegraphics}} command within an optional argument. (You can create
%  your own custom macro containing the $\backslash${\tt{includegraphics}} command to make things
%  simpler here.)
 
% \vspace{11pt}

 \vspace{-30pt}
\begin{IEEEbiography}[{\includegraphics[width=1in,height=1.25in,clip,keepaspectratio]{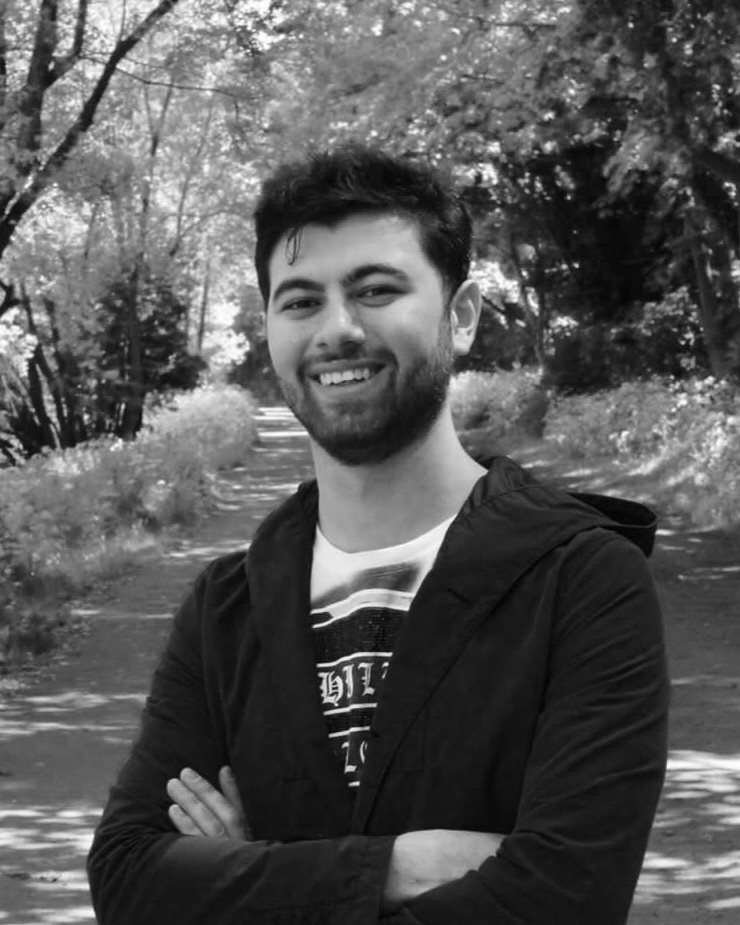}}]{Riad Ibadulla} (Member, IEEE) received his BEng in Computer Systems Engineering from City, University of London, and his MSc in Artificial Intelligence from the University of St Andrews. After completing his PhD titled High-Resolution Capabilities of Free-space Optical Neural Networks, which explored novel methods to optimise Convolutional Neural Networks and Vision Transformers for optical AI accelerators, he worked as a research assistant on the development of robust IDS by combining machine learning with formal methods. He is currently a lecturer in computer science at City St George’s, University of London, and his primary areas of expertise are deep learning and computer vision.
\end{IEEEbiography}

% \vspace{11pt}

 \vspace{-30pt}
\begin{IEEEbiography}[{\includegraphics[width=1in,height=1.25in,clip,keepaspectratio]{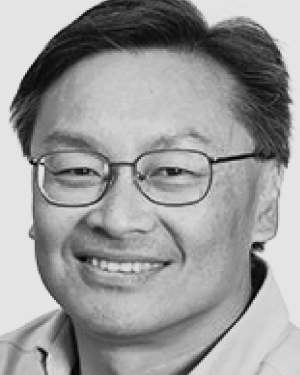}}]{Thomas M. Chen} (Senior Member, IEEE) received the B.S. and M.S. degrees in electrical engineering from the Massachusetts Institute of Technology, Cambridge, MA, USA, in 1983 and 1984, respectively, and the Ph.D. degree in electrical engineering from the University of California, Berkeley, CA, USA, in 1990., Recently he was a Professor in Cyber Security at the School of Engineering and Mathematical Sciences, City University London, London, U.K. Previously, he was a Professor in Networks at Swansea University, Swansea, U.K., and an Associate Professor at Southern Methodist University, Dallas, TX, USA.,Prof. Chen received the IEEE Communications Society Fred Ellersick Best Paper Award in 1996. He has been Editor-in-Chief for IEEE Communications Surveys (1996–1997), IEEE Communications Magazine (2006–2007), and IEEE Network (2009–2011). He served as an Associate Editor for the Journal on Security and Communication Networks and the International Journal of Digital Crime and Forensics.
\end{IEEEbiography}

% \vspace{11pt}
 \vspace{-31pt}
\begin{IEEEbiography}[{\includegraphics[width=1in,height=1.25in,clip,keepaspectratio]{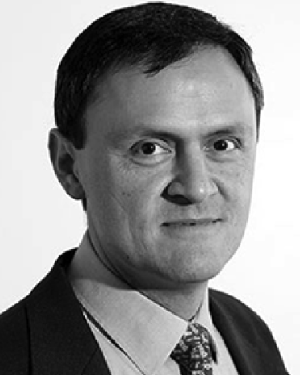}}]{Constantino Carlos Reyes-Aldasoro} (Senior Member, IEEE) received the B.Sc. degree in electrical engineering from Universidad Nacional Autónoma de México, in 1993, the M.Sc. degree in communications and signal processing from the Imperial College of Science Technology and Medicine, in 1994, and the Ph.D. degree in computer science from Warwick University, in 2005. He was a Postdoctoral Research Associate with the Tumor Microcirculation Group, Medical School, University of Sheffield, from 2005 to 2011. He was with Sussex University, from 2011 to 2013, and joined the City, University of London, in 2013. His research interests include image analysis of biomedical images with special interest in cancer, inflammation, and microcirculation. He has worked with data from x-rays, computed tomography, magnetic resonance imaging, light, fluorescent, confocal, multiphoton, and electron microscopy. He served as the Chair for the Vision and Imaging Technical Network of the Institute of Engineering and Technology and a member for Executive Committees of the British Association for Cancer Research and the Royal Microscopical Society. He was the Chair of the 2014 Medical Image Understanding and Analysis Conference and the 2022 British Machine Vision Conference.
\end{IEEEbiography}

% \bf{If you will not include a photo:}\vspace{-33pt}
% \begin{IEEEbiographynophoto}{John Doe}
% Use $\backslash${\tt{begin\{IEEEbiographynophoto\}}} and the author name as the argument followed by the biography text.
% \end{IEEEbiographynophoto}

\vfill

\end{document}